\documentclass{article}
\usepackage{ijcai21}

\usepackage{times}
\usepackage{soul}
\usepackage{url}
\usepackage[hidelinks]{hyperref}
\usepackage[utf8]{inputenc}
\usepackage[small]{caption}
\usepackage{graphicx}
\usepackage{amsmath}
\usepackage{amsthm}
\usepackage{amsfonts}
\usepackage{booktabs}
\usepackage{algorithm}
\usepackage{algorithmic}
\usepackage{subfigure}
\usepackage{color}
\usepackage{comment}

\title{Few-Shot Partial-Label Learning}

\author{
Yunfeng Zhao$^{1,2}$\and
Guoxian Yu$^{1,2}$\footnote{corresponding author}\and
Lei Liu$^{1,2}$\and
Zhongmin Yan$^{1,2}$\and
Lizhen Cui$^{1,2}$\and
Carlotta Domeniconi$^3$

\affiliations
$^1$School of Software Engineering, Shandong University, Jinan, Shandong, China\\
$^2$Joint SDU-NTU Centre for Artificial Intelligence Research, Shandong University, Jinan, China.\\
$^3$Department of Computer Science, George Mason University, VA, USA
\emails
yunfengzhao@mail.sdu.edu.cn, 
\{gxyu, l.liu, yzm, clz\}@sdu.edu.cn,
carlotta@cs.gmu.edu
}

\begin{document}

\maketitle

\begin{abstract}
Partial-label learning (PLL) generally focuses on inducing a noise-tolerant multi-class classifier by training on overly-annotated samples, each of which is annotated with a set of labels, but only one is the valid label. A basic promise of existing PLL solutions is that there are sufficient partial-label (PL) samples for training. However, it is more common than not to have just few PL samples at hand when dealing with new tasks. Furthermore, existing few-shot learning algorithms assume precise labels of the support set; as such,  irrelevant labels may seriously mislead the meta-learner  and thus lead to a compromised performance. How to enable PLL under a few-shot learning setting is an important problem, but not yet well studied. In this paper, we introduce an approach called FsPLL (Few-shot PLL).  FsPLL first performs adaptive distance metric learning by an embedding network and rectifying prototypes on the tasks previously encountered. Next, it calculates the prototype of each class of a new task in the embedding network. An unseen example can then be classified via its distance to each prototype. 
Experimental results on widely-used few-shot datasets (Omniglot and miniImageNet) demonstrate that our FsPLL can achieve a superior performance than the state-of-the-art methods across different settings, and it needs fewer samples for quickly adapting to new tasks.
\end{abstract}

\section{Introduction}

In partial label learning (PLL)~\cite{cour2011learning}, each `partial-label' (PL) training sample is annotated with a set of candidate labels, among which only one is the ground-truth label. The aim of PLL is to induce a noise-tolerant multi-class classifier from such PL samples. PLL is currently one of the most prevalent weakly-supervised learning paradigms, which include inaccurate supervision, where the given
labels do not always correspond to the ground-truth; 
incomplete supervision, where only a subset of the training data is labeled; and inexact supervision, where
the training data have only coarse-grained labels \cite{zhou2018brief}. This paper focuses on the first paradigm, where the given labels of the training data do not always represent the ground-truth. This learning problem arises in diverse domains, where a large number of inaccurately annotated samples can be easily collected, and it is very difficult (or impossible) to identify the true labels from the given ones \cite{zheng2017truth,tu2020jmf}.

Let $\mathcal{X} \in \mathbb{R} ^d $ denote the $d$-dimensional instance feature space and $\mathcal{Y}=\{0,1\}^l$ denote the label space with $l$ distinct labels. The aim of PLL is to learn a noise-robust multi-class classification model $f: \mathcal{X} \to \mathcal{Y}$ with  the PL dataset $\mathcal{D}=\{(\mathbf{x}_i,\mathbf{y}_i)|1\le i\le n\}$, where $\mathbf{x}_i \in \mathcal{X}$ is the feature vector of the $i$-th instance,   $\mathbf{y}_i$ is the one-hot label vector of candidate labels ($\mathcal{Y}_i\subset \mathcal{Y}$) of the $i$-th instance, and $z_i \in \mathcal{Y}_i$ is the unknown ground-truth label of this instance. The key challenge to address the PLL problem is to recover the ground-truth label concealed within the candidate label set for every training instance. Existing PLL methods can be roughly categorized into averaging-based disambiguation  and identification-based disambiguation. The former class of methods typically equally treats each candidate label during the process of model induction, and performs label prediction by averaging the modeling outputs~\cite{cour2011learning,gong2017regularization}. The second category of methods models the ground-truth label of the training instance as a latent variable, and estimates it via an iterative refining procedure~\cite{yu2017maximum,chai2019large,yu2018fPML}. 

\begin{figure}[tb]
\centering
\includegraphics[width=8cm, height=4cm]{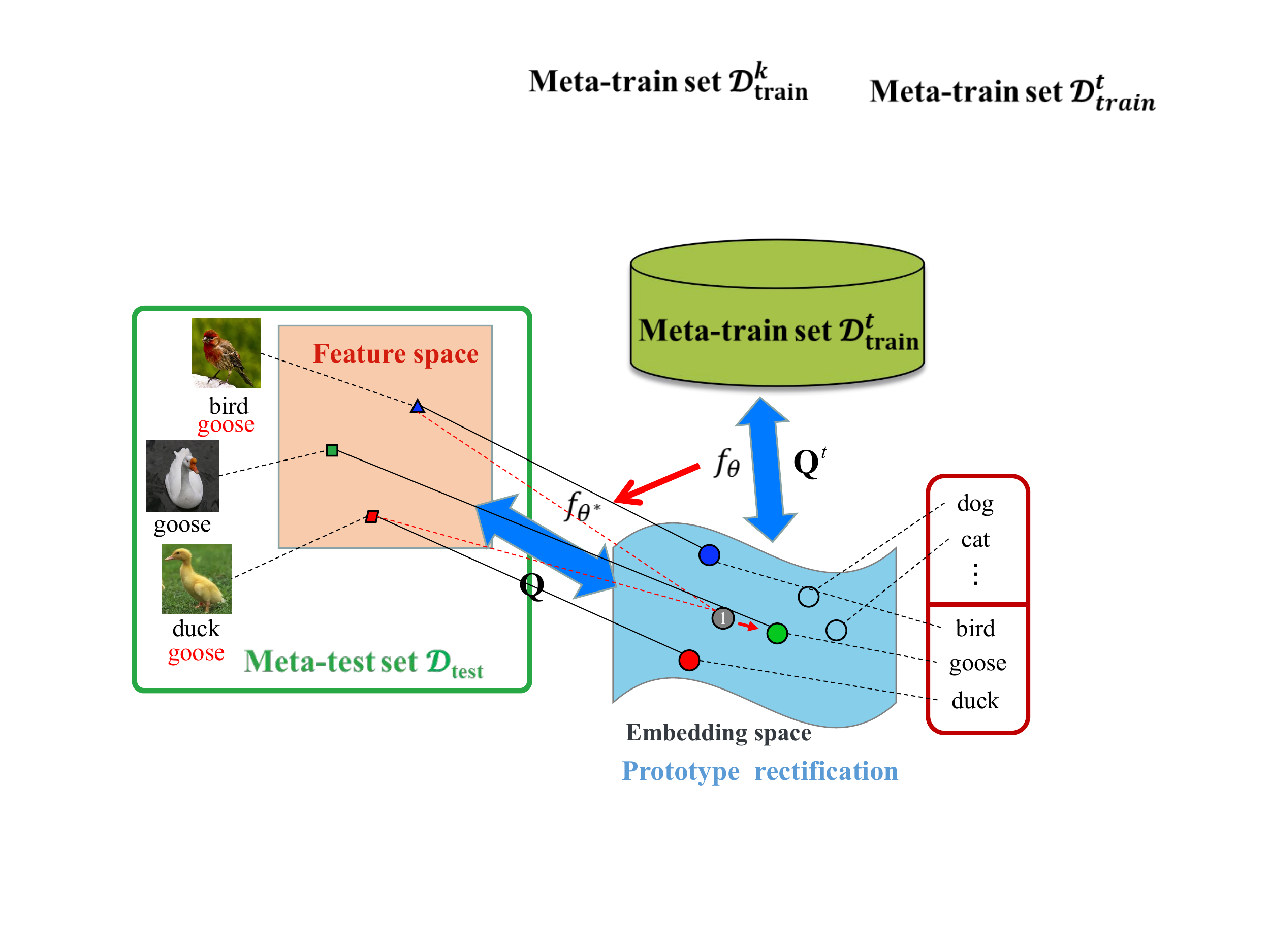}
\vspace{-1em}
 \caption{The overall schematic framework of FsPLL. FsPLL learns an embedding network ($f_\theta$) and rectifies label confidence matrix $\{\mathbf{Q}^t\}_{t=1}^T$ to perform adaptive distance metric learning based on PL samples previous encountered ($\mathcal{D}^t_{train}$). Next, it updates label confidence matrix $\mathbf{Q}$ and rectifies prototypes  w.r.t. the new task ($\mathcal{D}_{test}$) in the embedding space. An unseen example can then be classified via its distance to prototypes. The `circle with 1' in the embedding space is the contaminated prototype (no rectification) of `goose'.}
 \label{framework}
\end{figure}

These PLL approaches rely on the assumption that sufficient labeled/unlabeled training data which are relevant to the task are available. They don't perform well in a \textbf{few-shot} scenario, where each class has only few training samples, annotated with inaccurate labels. Although Few-Shot Learning (FSL) has been extensively applied in diverse domains \cite{snell2017prototypical,finn2017model,vanschoren2018meta}, the existing FSL methods typically assume that the labels of the few-shot support samples are \textbf{noise free}. Unfortunately, the violation of this assumption  seriously compromise the performance of the few-shot classifier, as shown in our experiments. To the best of our knowledge, how to make FSL effective with few-shot PL samples, is an open and under-studied problem. To bridge this gap, we propose a Few-shot PLL approach (\textbf{FsPLL}), which is based on the prototypical network~\cite{snell2017prototypical} {and on the local manifold~\cite{belkin2006manifold} in feature space, which states that instances that have similar feature vectors are more likely to share a same ground-truth label.
More specifically, FsPLL first aims at iterative rectifying the ground-truth class prototypes of support PL samples and learning an embedding network, where, based on the previous tasks, every sample is closer to its ground-truth prototype, and further apart from its non-ground-truth prototypes.} Next, it calculates the prototype of each class of the new task  by the embedding network and prototype rectification. Then, an unseen example can be  classified via its distance to each class prototype. The whole framework of FsPLL is illustrated in Fig. \ref{framework}.

The main contributions of our work are as follows:\\
\noindent (i) We focus on a practical and general PLL setting, where the training samples of the target task are few-shot. We also tackle the  problem of noisy labels of few-shot support samples, which can seriously mislead the meta-learner when adapting to the target task. Both issues are not addressed by existing PLL solutions and few-shot/meta learning methods.\\
\noindent (ii) {We introduce a prototype rectification strategy with prototypical embedding network  to learn the underlying ground-truth prototypes of  support and query PL samples, which is less impacted by irrelevant labels and can more credibly adapt to new tasks.}\\
\noindent (iii)  Extensive experiments on  benchmark few-shot datasets show that our FsPLL outperforms the state-of-the-art PLL approaches \cite{zhang2017disambiguation,zhang2016partial,wu2018towards,wang2019adaptive} and baseline FSL methods \cite{snell2017prototypical,finn2017model}. The overlook of irrelevant labels of few-shot PL samples indeed seriously compromises the performance of FSL methods, and our FsPLL can greatly remedy this problem.


\section{Related work}
\label{sec:relwork}
\subsection{Partial Label Learning }
PLL is different from learning from noisy labels \cite{natarajan2013learning}, where  training samples are incorrectly annotated with the wrong label; it is also different from semi-supervised learning \cite{belkin2006manifold}, where some training samples are completely unlabeled but can be leveraged for training; and also different from weak-label learning \cite{sun2010multi,dong2018learning}, where the labels of training samples are incomplete. The current efforts for PLL can be roughly grouped into two categories: the averaging-based and the identification-based disambiguation.

The averaging-based disambiguation technique generally induces the classifier model by treating all candidate labels equally. Following this protocol, some instances-based methods~\cite{hullermeier2006learning,gong2017regularization} classify the ground-truth $y$ of an unseen instance $\mathbf{x}$ by averaging the candidate labels of its neighbors, i.e., $y=\arg \max _{y \in \mathcal{Y}} \sum_{\mathbf{x}_{i} \in \mathcal{N}(\mathbf{x})} \mathbb{I}\left(y \in \mathcal{S}_{i}\right)$, where $\mathcal{S}_{i}$ denotes the candidate label set  of the $i$-th instance and $\mathcal{N}(\mathbf{x})$ denotes the set of neighbors of instance $\mathbf{x}$, while other parametric methods~\cite{cour2011learning,zhang2016partial} aim at inducing a parametric model $\theta$ by maximizing the gap between the average modeling output of  the candidate labels and that of the non-candidate ones, i.e., $\max (\sum_{i=1}^{m}(\frac{1}{|\mathcal{S}_{i}|} \sum_{y \in \mathcal{S}_{i}} F(\mathbf{x}_{i}, y; \theta) - \frac{1}{|\widehat{\mathcal{S}}_{i}|} \sum_{\widehat{y} \in \widehat{\mathcal{S}}_{i}} F(\mathrm{x}_{i}, \widehat{y} ;\theta)))$ where $\widehat{\mathcal{S}}_{i}$  denotes the set of non-candidate labels. As to the identification-based disambiguation technique, the ground-truth labels of the training instances are seen as latent variables and to be optimized by an iterative refining procedure. Following this paradigm, some methods train the model based on the maximum likelihood criterion~\cite{jin2002learning} or the maximum margin criterion~\cite{nguyen2008classification}. Recently, some teams mine the topological information~\cite{zhang2016partial,feng2018leveraging} in the instance feature space to help the optimization of label confidence. 

Nevertheless, although these methods can disambiguate labels and induce a noise-tolerance classifier by different techniques,  they can hardly work in a more universal scenario, in which the PL samples we collected are few-shot, which break the premise of many-shot training samples per label for inducing a PLL classifier. In fact, existing PLL methods still work in a close label set fashion. But in practice, we may often come into new scenarios, where we can only collect few-shot PL samples and each target label is annotated to several samples. To enable PLL in this general setting, we propose FsPLL to learn noise-robust class prototypes by an embedding network and by rectifying prototypes therein. 



\subsection{Few-shot Learning}
Few-Shot Learning (FSL) \cite{fei2006one} is an example of meta-learning \cite{huisman2020survey}, where a learner is trained on several related tasks during the meta-training phase, so that it can generalize well to unseen (but related) tasks using just few samples with supervision during the meta-testing phase. Existing FSL solutions mainly focus on supervised learning problems, and usually one may term as \emph{$N$-way $K$-shot} classification, where $N$ stands for the number of classes and $K$ means the number of training samples per class, so each task contains $KN$ samples. Given limited support samples for training, unreliable empirical risk minimization is the core issue of FSL, and existing solutions for FSL can be grouped from the perspective of data, model and algorithm \cite{wang2020generalizing}. Data augmentation-based FSL methods aim to acquire more supervised training samples by generating more samples from original few-shot samples, weakly-labeled/unlabeled data or similar datasets \cite{douze2018low}, and thus to reduce the uncertainty of empirical risk minimization. Model-based FSL methods typically manage to shrink the ambient hypothesis space into a smaller one by extracting prior knowledge in the meta-training phase \cite{snell2017prototypical,sung2018learning}, so the empirical risk minimization becomes more reliable and the over-fitting issue is reduced. Algorithm-based FSL approaches use prior knowledge to guide the seek of optimal model parameters by providing a good initialized parameter  or directly learning an optimizer for new tasks \cite{finn2017model}.


Unfortunately, most FSL methods ideally assume the support samples in meta-testing set is with accurate supervision, namely, these samples are precisely annotated with labels. But these support samples are PL ones with irrelevant labels, which mislead the adaption of FSL methods toward the target task (as shown in Fig. \ref{framework}) and cause a compromised performance. {To address this problem, our FsPLL performs the optimization of embedding network and prototype rectification therein in an iterative manner. In this way, the learnt embedding network and prototypes are less impacted by irrelevant labels of PL samples, and can credibly adapt to new tasks.} 



\section{The Proposed Methodology}
\label{sec:method}
Suppose we are given a small support/training set of $n$ PL samples  $\mathcal{D}=\{(\mathbf{x}_i,\mathbf{y}_i)|1\le i\le n\}$ and its corresponding label space and feature space are $\mathcal{Y}=\{0,1\}^l$ and $\mathcal{X} \in \mathbb{R}^d $, respectively. The goal of FsPLL is to induce a multi-class classifier $f: \mathcal{X} \to \mathcal{Y}$, which can precisely predict the ground-truth label of an unseen instance $\mathbf{x}$ under this few-shot classification scenario. Different from existing PLL methods, FsPLL should and can utilize the knowledge previously acquired from meta-training phase to quickly adapt to the new classification task $\mathcal{D}$ in the meta-testing phase. In the meta-training phase, FsPLL learns an embedding network (meta-knowledge) to project PL samples more nearby with their ground-truth prototypes and apart from their non ground-truth prototypes by iteratively rectifying these prototypes in this embedding space. In the meta-test phase,  it rectifies the prototypes of support PL samples using the embedding work and then classifies new PL samples by their distance to rectified prototypes in the embedding space. In this paper, we take Prototypical Network (PN) \cite{snell2017prototypical} as the base of our embedding network. The framework overview of FsPLL is given in Fig. \ref{prototpye}.  The following subsections elaborates on the two phases. 

\subsection{Meta-training phase}
The meta-training phase mainly aims to extract prior knowledge from multiple relevant tasks for the target task. Suppose we are given $T \gg 1$ few-shot datasets (tasks) denoted as $\mathcal{D}_{train}^{t}~(1\le t\le T)$. For each dataset  $\mathcal{D}_{train}^t=\{\mathbf{X}^t_s,\tilde{\mathbf{X}}^t_q,\mathbf{Y}^t\}$, where $\mathbf{X}^t_s= (\mathbf{x}^t_1, \mathbf{x}^t_2,\dots ,\mathbf{x}^t_{n_s}) \in\mathbb{R}^{d\times n_s}$ denotes the data matrix of support samples, $\tilde{\mathbf{X}}^t_q = (\tilde{\mathbf{x}}^t_1, \tilde{\mathbf{x}}^t_2, \cdots, \tilde{\mathbf{x}}^t_{n_q})\in\mathbb{R}^{d\times n_q}$ denotes data matrix of query samples, $\mathbf{Y}^t=(\mathbf{y}^t_1,\mathbf{y}^t_2,\cdots ,\mathbf{y}^t_{n_s})\in\mathbb{R}^{l\times {n_s}}$ is the corresponding label matrix of support samples, and $n_s+n_q < n$. $\mathbf{Y}^t_{ci} = 1$ means the $c$-th label is a candidate label of the $i$-th sample; $\mathbf{Y}^t_{ci} = 0$ otherwise.
Let $\mathbf{Q}^t\in \mathbb{R}^{l\times n_s}$ denotes the underlying label confidence matrix of support samples and it is initialized as $\mathbf{Y}^t$, where $\mathbf{Q}^t_{ci}$ indicates the confidence of the $c$-th label as the ground-truth label of the $i$-th sample. 

From these datasets, we aim at learning an embedding network, i.e., $f_{\theta}:\mathbb{R}^d\to\mathbb{R}^m$, by which we can  obtain the representation of every label in the embedding space and can be more robust to irrelevant labels of support samples therein.
Suppose $\mathbf{P}^t = (\mathbf{p}^t_1,\mathbf{p}^t_2,\dots,\mathbf{p}^t_l) \in \mathbb{R}^{l\times m}$ is the prototype/representation matrix of $l$ class labels of the $t$-th task, where $\mathbf{p}^t_c$ denotes the prototype of the $c$-th label in the embedding space.
PN~\cite{snell2017prototypical} computes the prototype by $\mathbf{p}^t_c =\frac{\sum_{i=1}^{n_s}\mathbf{Y}^t_{ci}\times f_{\theta}(\mathbf{x}^t_i) }{\sum_{i=1}^{n_s}\mathbf{Y}^t_{ci}}$, which equally treats all PL samples annotated with $c$ to induce the prototype, neglects that some PL samples actually not annotated with this label. Therefore, PN gives contaminated prototypes. {For example,  prototype of goose (`circle with 1') in Fig. \ref{framework} is 
misled by irrelevant labels}.  These prototypes consequently compromise the classification performance, especially when support PL samples with excessive irrelevant labels. To address this issue, FsPLL performs prototype rectification and label confidence update in an iterative way to seek noise-robust embedding network and prototypes in the embedding space, as shown in Fig. \ref{framework}. FsPLL defines each prototype based on the confidence weighted mean of the corresponding support samples in the embedding space as follows:
\begin{equation}
\label{prototpye}    
\mathbf{p}_c = \frac{\sum_{i=1}^{n_s}\mathbf{Q}_{ci}\times f_{\theta}(\mathbf{x}_i) }{\sum_{i=1}^{n_s}\mathbf{Q}_{ci}},
\end{equation}
Unlike the prototypes optimized by PN, FsPLL rectifies the prototypes using iterative updated label confident matrix $\mathbf{Q}^t$, and thus explicitly accounts for the irrelevant labels of training samples.  

It is expected for a sample to be closer to its ground-truth prototype in the embedding space; this would enable a confident label prediction in this space. Given this, we use a softmax to update the label confidence matrix  $\mathbf{Q}^t$ as follows: 
\begin{equation}
\label{confidence1}
\mathbf{Q}_{ci}= 
\begin{cases}
   \frac{\text{exp}(-d(f_{\theta}(\mathbf{x}_i),\mathbf{p}_c))}{{\sum}_{c=1}^{l}\text{exp}(-d(f_{\theta}(\mathbf{x}_i),\mathbf{p}_c))\times\mathbf{Y}_{ci}} & \text{ if } \mathbf{Y}_{ci}=1 \\ \text{~~~~~~~~~~~~~~~~~~~~~~~0}
  & \text{ otherwise } 
\end{cases},
\end{equation}
where $d(f_{\theta}(\mathbf{x}^t_i),\mathbf{p}^t_c)$ quantifies the Euclidean distance between sample $\mathbf{x}^t_i$ and prototype $\mathbf{p}^t_c$ in the embedding space. The labels of a PL sample can be disambiguated by referring to labels of its neighborhood samples \cite{zhang2016partial,wang2019adaptive}. We observe that PN and Eq. \eqref{prototpye} disregard the neighborhood support samples when computing the prototype. Unlike these PLL methods that disambiguate in the original feature space or linearly projected subspace, FsPLL further updates the label confidence matrix in the embedding space as follows:  
\begin{equation}
\label{confidence2}
\mathbf{Q}_{ci}= 
   \mathbf{Q}_{ci}+\frac{\lambda}{|\mathcal{N}_k(\mathbf{x}_i)|}\sum_{\mathbf{x}_j \in \mathcal{N}_k(\mathbf{x}_i)}\mathbf{Q}_{cj},\ \text{ if } \mathbf{Y}_{ci}=1 
\end{equation}
where $\mathcal{N}_k(\mathbf{x}_i^t)$ includes the $k$-nearest samples of $\mathbf{x}_i^t$, and the neighborhood is determined by Euclidean distance in the embedding space. $\lambda$ trade-offs the confidence from the sample itself and those from neighborhood samples. In this way, FsPLL utilizes local manifold of samples to rectify prototypes.


Based on the rectified prototypes and embedding network $f_{\theta}$, we can predict the label of a query sample with a softmax over its distances to all prototypes in the embedding space as:
\begin{equation}
    \label{loss}
    p_{\theta}(z^t_j = c~|~\tilde{\mathbf{x}}^t_j) = \frac{\text{exp}(-d(f_{\theta}(\tilde{\mathbf{x}}^t_j),\mathbf{p}^t_c))}{\sum_{i=1}^{l}\text{exp}(-d(f_{\theta}(\tilde{\mathbf{x}}^t_j),\mathbf{p}^t_i))},
\end{equation}
where $z^t_j$ is the unknown ground-truth label of the $j$-th query sample. To make the representation of every query sample in the embedding space closer to its ground-truth prototype and apart from its non ground-truth prototypes, FsPLL minimizes the negative log-probability of the most likely label of a query example as follows:
\begin{equation}
\mathbf{J}(\theta,\tilde{\mathbf{x}}^t_i)=-\log (\max _{c=1,\cdots,l }p_\theta(z^t_j = c~|~\tilde{\mathbf{x}}^t_i))
\label{negativeLoss}
\end{equation}
By minimizing the above equation, FsPLL can obtain the  rectified prototypes $\mathbf{P}^t$ and the corresponding embedding network parameterized by $f_{\theta}$ for task $\mathcal{D}_{train}^t$. We want to remark that the $c$-th label for different tasks is not always the same.

The meta-training phase involves  a lot of different tasks, each of which is composed of support/query samples. To enable a good generalization ability, it attempts to gain the optimal mode parameter $\theta^*$ by minimizing the average negative log-probability of the most likely labels of all query samples over $T$ tasks as follows:  
\begin{equation}
\label{theat}
    \theta^* = {\arg\min}_{\theta } {\sum}_{t=1}^T \frac{1}{n_q}{\sum}_{i=1}^{n_q} \mathbf{J}(\theta,\tilde{\mathbf{x}}^t_i)
\end{equation}
To this end, FsPLL obtains an embedding network $f_{\theta^*}$ that is robust to irrelevant labels of PL samples across $T$ tasks. Via this network, a PL sample in the embedding space is made closer to its ground-truth prototype than to other prototypes, and the generalization and fast adaption ability are pursued among $T$ different tasks. 

\subsection{Meta-test phase}
In the meta-test phase, we are only given a small set of PL samples, which compose the target task with support and query samples. These support samples are overly-annotated with irrelevant labels, while query samples are without label information. We want to highlight that the labels of these PL samples are \emph{disjoint} with the labels used in the meta-training phase. In other words, the PL samples are few-shot ones. Here, FsPLL aims to use the knowledge (embedding network $f_{\theta^*}$) acquired in the meta-training phase to precisely annotate the query samples based on the inaccurately supervised few-shot support examples.

Formally,  FsPLL aims to quickly generalize to a new task $\mathcal{D}_{test}=\{\mathbf{X}_s,\tilde{\mathbf{X}}_q,\mathbf{Y}\}$, where $\mathbf{X}_s\in\mathbb{R}^{d\times n_s}$, $\tilde{\mathbf{X}}_q\in\mathbb{R}^{d\times n_q}$ and $\mathbf{Y}\in\mathbb{R}^{l\times n_s}$ denote the data matrices of support examples, of query examples, and of labels of query examples, respectively. Alike the meta-training phase, FsPLL first computes the prototypes $\mathbf{P}\in \mathbb{R}^{m\times l}$ of this new task in the embedding space using the confidence-weighted mean of support samples $\mathbf{X}_s$ and label confidence matrix  $\mathbf{Q}$ as in Eq. \eqref{prototpye}. Then the label confidence matrix  $\mathbf{Q}$ of the support samples is updated based on a softmax over their distances to prototypes as in Eq. \eqref{confidence2}. FsPLL repeats the above two steps to  rectify the prototypes and update label confidence matrix for adapting to the target task. Note, the embedding network $f_{{\theta}^*}$ is fixed during the above repetitive optimization. 

Given a query sample $\mathbf{x}_i$, FsPLL classifies it  using  its distance to rectified prototypes $\mathbf{P}\in \mathbb{R}^{m\times l}$ as follows:
\begin{equation}
\label{classify}
    z_i = \arg\max_q{p_{\theta^*}(z_i=  q~|~\mathbf{x}_i)~~(q = 1,\cdots,l)}
\end{equation}
Algorithm~\ref{alg:Model} lists the procedure of 
meat-training phase (step 1-6) and meta-test phase (step 7-10) of FsPLL.

\begin{algorithm}[bt]
\caption{FsPLL: Few-shot Partial Label Learning}
\label{alg:Model}

\textbf{Input}: $\{\mathcal{D}_{train}^{t}\}_{t=1}^T$, $\mathcal{D}_{test}$, $maxEpoch=200$ (number of epoches for training embedding network), $\lambda=0.5$, $k=K_2-1$.\\
\textbf{Output}:  Embedding model $f_{\theta^*}$, predicted labels of query samples $\{z_1,z_2,\cdots,z_{n_q}\}$
\label{Model}
\vspace{-0.3em}
\begin{algorithmic}[1] 
        \FOR {$epoch = 1\to maxEpoch$}
            \FOR {$ t = 1\to T$}
                \STATE Update   $\mathbf{P}^t\in \mathbb{R}^{m\times l}$, $\mathbf{Q}^t\in \mathbb{R}^{l\times n_s}$ via Eq. \eqref{prototpye} and Eq. \eqref{confidence2} in an iterative way. 
            \ENDFOR
            \STATE Calculate loss via Eq. \eqref{theat} and updata $\theta$ via SGD.
        \ENDFOR
        \STATE Update   $\mathbf{P}\in \mathbb{R}^{m\times l}$, $\mathbf{Q}\in \mathbb{R}^{l\times n_s}$ via Eq. \eqref{prototpye} and Eq. \eqref{confidence2} using  $f_{\theta^*}$, $\mathbf{X}_s \in \mathcal{D}_{test}$ and $\mathbf{Y}$ in an iterative way. 
        \FOR{$i = 1 \to n_q$}
            \STATE Get $z_i$ via Eq. \eqref{classify}
        \ENDFOR
\end{algorithmic}
\end{algorithm}

\section{Experiments}
\label{sec:exp}

\subsection{Experimental Setup}
\textbf{Datasets}: We conduct experiments on two benchmark FSL datasets (\textbf{Omniglot}~\cite{lake2011one}  and \textbf{miniImageNet}~\cite{vinyals2016matching}). Following the canonical protocol adopted by previous PLL methods \cite{wang2019adaptive,zhang2016partial}, we generate the semi-synthetic PL datasets on Omniglot and miniImageNet by two controlling parameters $p$ and $r$. $p$ controls the proportion of PL samples, and $r$ controls the number of irrelevant labels of a PL sample. Each $\mathcal{D}_{train}^t$ consisted of $N_1=30$ classes were randomly sampled from 4800/80 classes of Omniglot/miniImageNet without replacement. As to the meta-test set, we randomly selected another $N_2$ classes from 1692/20 test classes without replacement. For each selected class, $K_1=5$ ($K_2$)  samples were randomly chosen from 20/600 samples without replacement for the meta-training (meta-test) support samples, and the remaining/15 samples per class were randomly chosen as the query samples. More details on data split are given in the Supplementary file.

\begin{table*}[tb]
	\centering
	\vspace{-2em}      
\begin{tabular}{c|ll|ll|ll|ll|}		
		\multicolumn{1}{l|}{} & \multicolumn{2}{c|}{$N_2=5$} & \multicolumn{2}{c|}{$N_2=10$} & \multicolumn{2}{c|}{$N_2=20$}  & \multicolumn{2}{c|}{$N_2=30$}\\
		\hline
		\multicolumn{1}{l|}{} & \multicolumn{1}{c}{$K_2=5$} & \multicolumn{1}{c|}{$K_2=10$} & \multicolumn{1}{c}{$K_2=5$} & \multicolumn{1}{c|}{$K_2=10$} & \multicolumn{1}{c}{$K_2=5$} & \multicolumn{1}{c|}{$K_2=10$} & \multicolumn{1}{c}{$K_2=5$} & \multicolumn{1}{c|}{$K_2=10$}\\
		\hline
	&	\multicolumn{8}{c|}{$r=1$} \\
		\hline  $\text{FsPLL}$ &\textbf{.892±.083} & \textbf{.895±.051}&\textbf{.789±.045}    & \textbf{.823±.067} & \textbf{.712±.034} & \textbf{.757±.056} & \textbf{.665±.015}  &\textbf{.701±.046}
		\\
		FsPLL-nM 
   & .852±.092          & .886±.072          & .776±.070          & .816±.062          & .695±.053          & .745±.047     &.643±.008   & .693±.042    \\
 
$\text{PN}$       & .579±.104          & .636±.105          & .435±.070          & .485±.071          & .317±.043          & .360±.044     &  .255±.032 &  .291±.034   \\

$\text{MAML}$     &  .673±.079&  .647±.097&  .592±.067&  .642±.053&  .514±.061& .544±.032 & .421±.018&.475±.065 \\

PL-AGGD   & .664±.118          & .777±.103          & .576±.086          & .714±.076          & .450±.053          & .649±.063   &.459±.043   & .601±.057      \\
PALOC     & .616±.116          & .726±.111          & .528±.075          & .651±.078          & .447±.058          & .568±.058   & .392±.047  & .513±.049     \\
PL-ECOC    & .473±.065          & .694±.109          & .513±.076          & .619±.073          & .279±.040          & .418±.054 &.363±.042 &.465±.046        \\
PL-LEAF    & .629±.117          & .768±.107          & .568±.087          & .712±.070          & .495±.060          & .630±.060   &.452±.043   & .592±.054      \\\hline
 $\text{FsPLL}^{+}$     & \textbf{.995±.029}          & \textbf{.997±.009}          & \textbf{.990±.020}          & \textbf{.993±.012}          & \textbf{.986±.009}          & \textbf{.990±.010}       &  \textbf{.981±.009} & \textbf{.986±.008} \\
 $\text{PN}^{+}$       & .965±.046          & .985±.030          & .956±.037          & .981±.021          & .939±.027          & .968±.020      &  .924±.025 &  .958±.018  \\
 $\text{MAML}^{+}$       & .858±.016& .902±.036 &  .849±.026&  .877±.014& .774±.019 & .831±.023 &  .638±.035 & .795±.189\\
 \hline 
          & \multicolumn{8}{c|}{$r=2$}                                                                                          \\\hline $\text{FsPLL}$&\textbf{.673±.098}  & \textbf{.742±.073}  &    \textbf{.712±.068}   &\textbf{.756±.063}   &\textbf{.654±.049}  &  \textbf{.689±.063}&\textbf{.598±.063}&.\textbf{602±.038}\\
          FsPLL-nM 
     & .616±.171          & .706±.136          & .566±.100          & .665±.087          & .494±.065          & .584±.056&.442±.053& .527±.048         \\
     
$\text{PN}$       & .476±.121          & .559±.114          & .378±.073          & .442±.074          & .270±.044          & .321±.047       &.213±.032 & .258±.033  \\

$\text{MAML}$       & .476±.121&  .553±.098& .458±.078& .549±.093 & .427±.037 & .472±.075 &.397±.043& .437±.036\\

PL-AGGD   & .496±.131          & .668±.129          & .490±.086          & .664±.082          & .451±.055          & .578±.061         &.416±.053&.545±.054 \\
PALOC     & .473±.117          & .611±.125          & .456±.083          & .591±.086          & .385±.056          & .525±.057 &.402±.049&.524±.061          \\
PL-ECOC    & .425±.105          & .481±.114          & .361±.081          & .531±.086          & .175±.044          & .343±.047          &.292±.046&.408±.052\\
PL-LEAF    & .484±.134          & .650±.125          & .489±.086          & .645±.083          & .436±.059          & .586±.060         &.398±.045&.525±.058 \\\hline
$\text{FsPLL}^{+}$ 
    & \textbf{.975±.076}         & \textbf{.997±.009}         & \textbf{.991±.010}          & \textbf{.994±.010}          & \textbf{.986±.009}          & \textbf{.989±.012}   &\textbf{.980±.009}&\textbf{.985±.007}       \\
    $\text{PN}^{+}$       & .825±.117          & .926±.076          & .871±.064          & .945±.041          & .850±.047          & .926±.032      &.826±.040&.908±.030
    \\
    $\text{MAML}^{+}$      &.675±.076&  .798±.056&  .783±.024& .760±.076 &  .668±.023&    .727±.025&.619±.026&.679±.450 \\

      \hline     
\end{tabular}
\vspace{-0.5em}
	\caption{Classification accuracy (mean±std) of comparison methods on \textbf{Omniglot}. \{FsPLL, PN, MAML\}$^+$ separately use the precise labels of meta-training support samples. $N_2$: the number of support classes; $K_2$: the number of training samples per class. The best performance in each  setting is \textbf{boldface}.}
\label{omniglot}
\end{table*}

\begin{figure*}[h!tbp]
\centering
\subfigure[$N_2=5$]{\includegraphics[width=4.2cm, height=3cm]{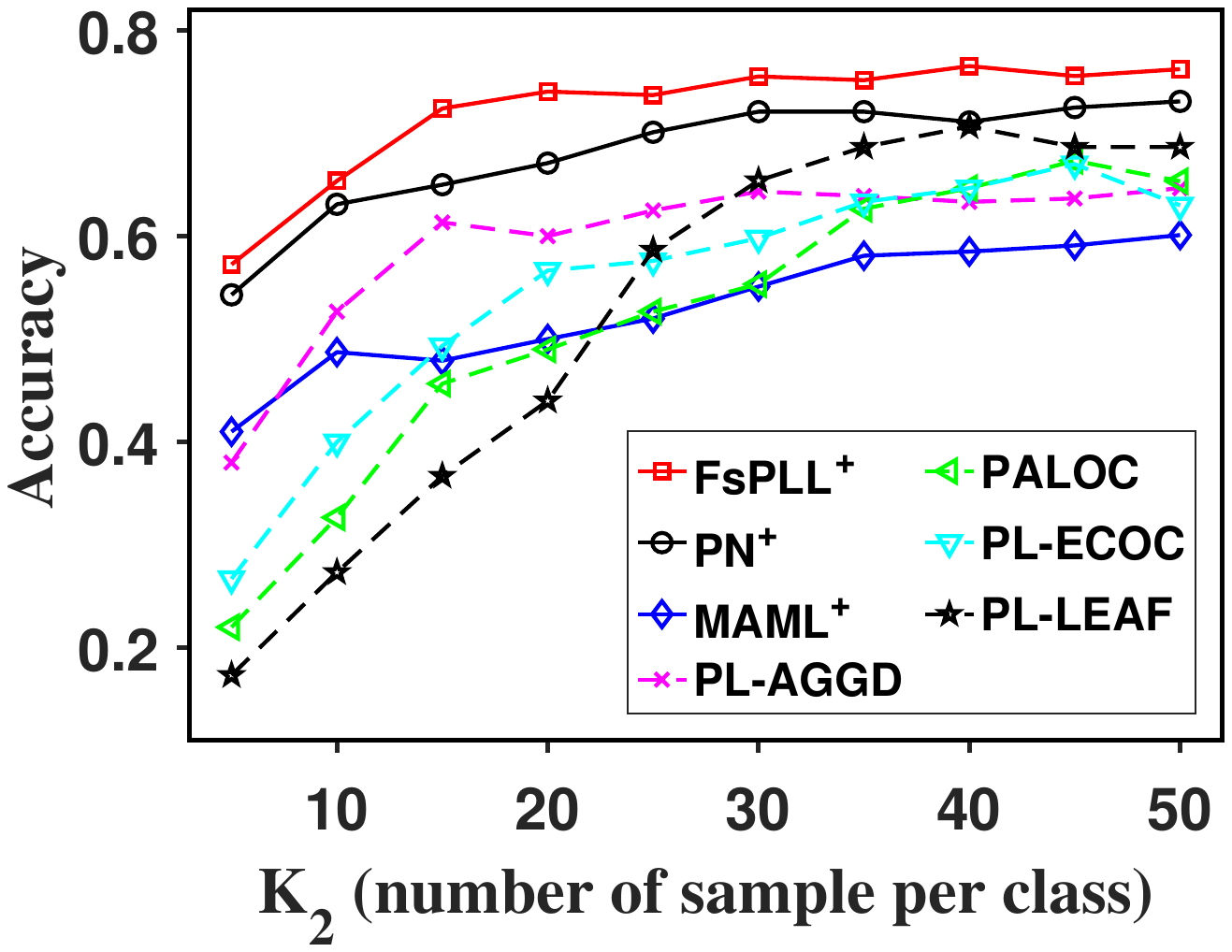}}
\subfigure[$N_2=10$]{\includegraphics[width=4.2cm, height=3cm]{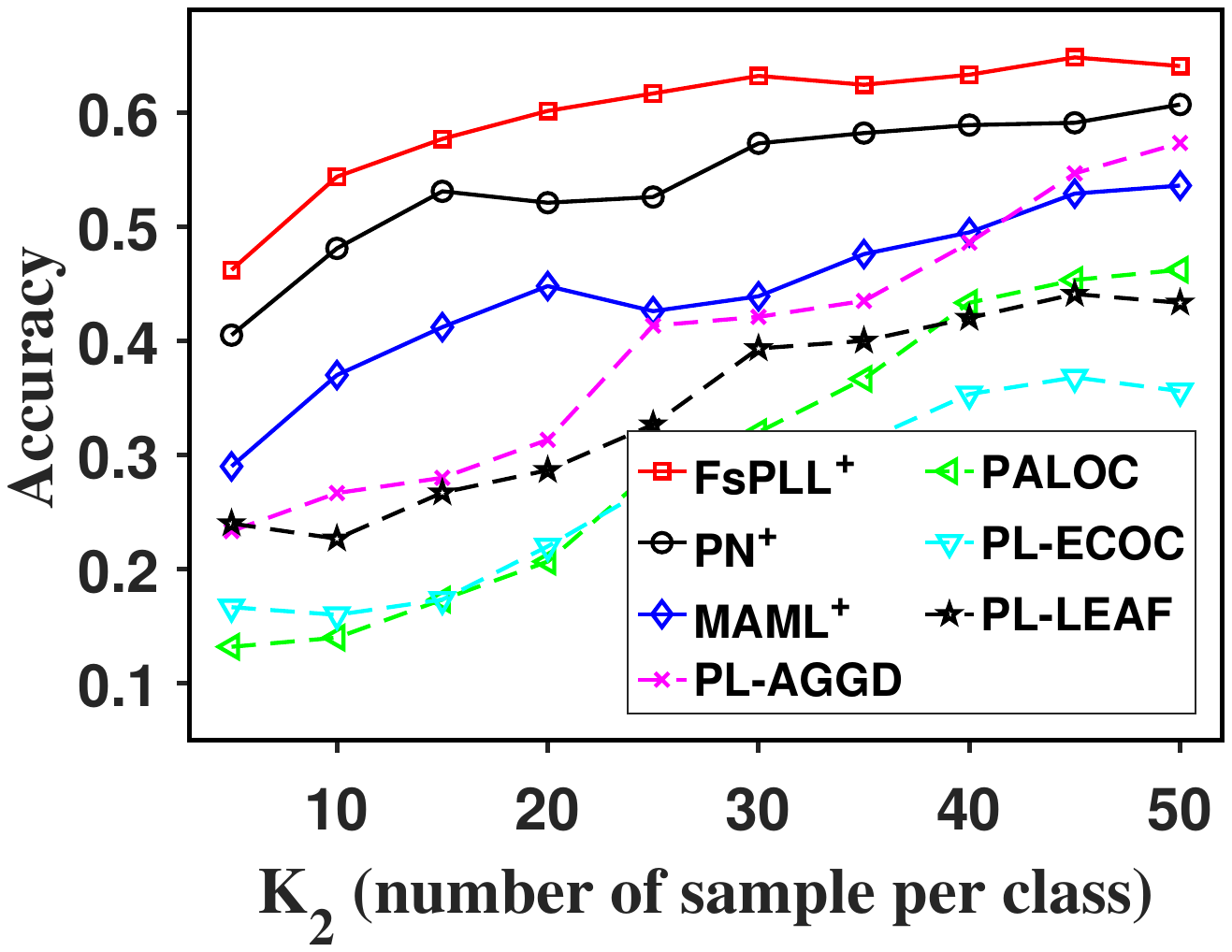}}
\subfigure[$N_2=15$]{\includegraphics[width=4.2cm, height=3cm]{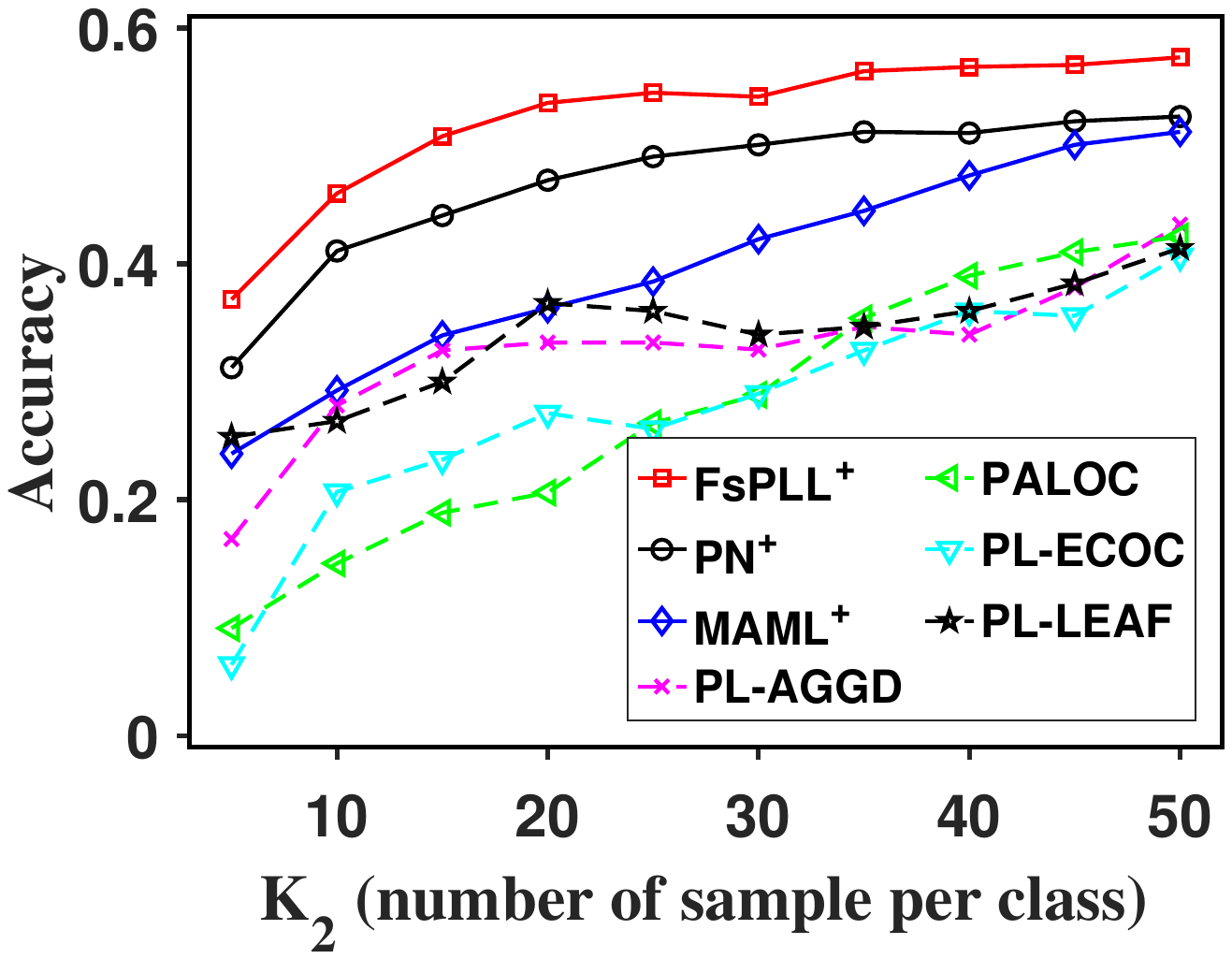}}
\subfigure[$N_2=20$]{\includegraphics[width=4.2cm, height=3cm]{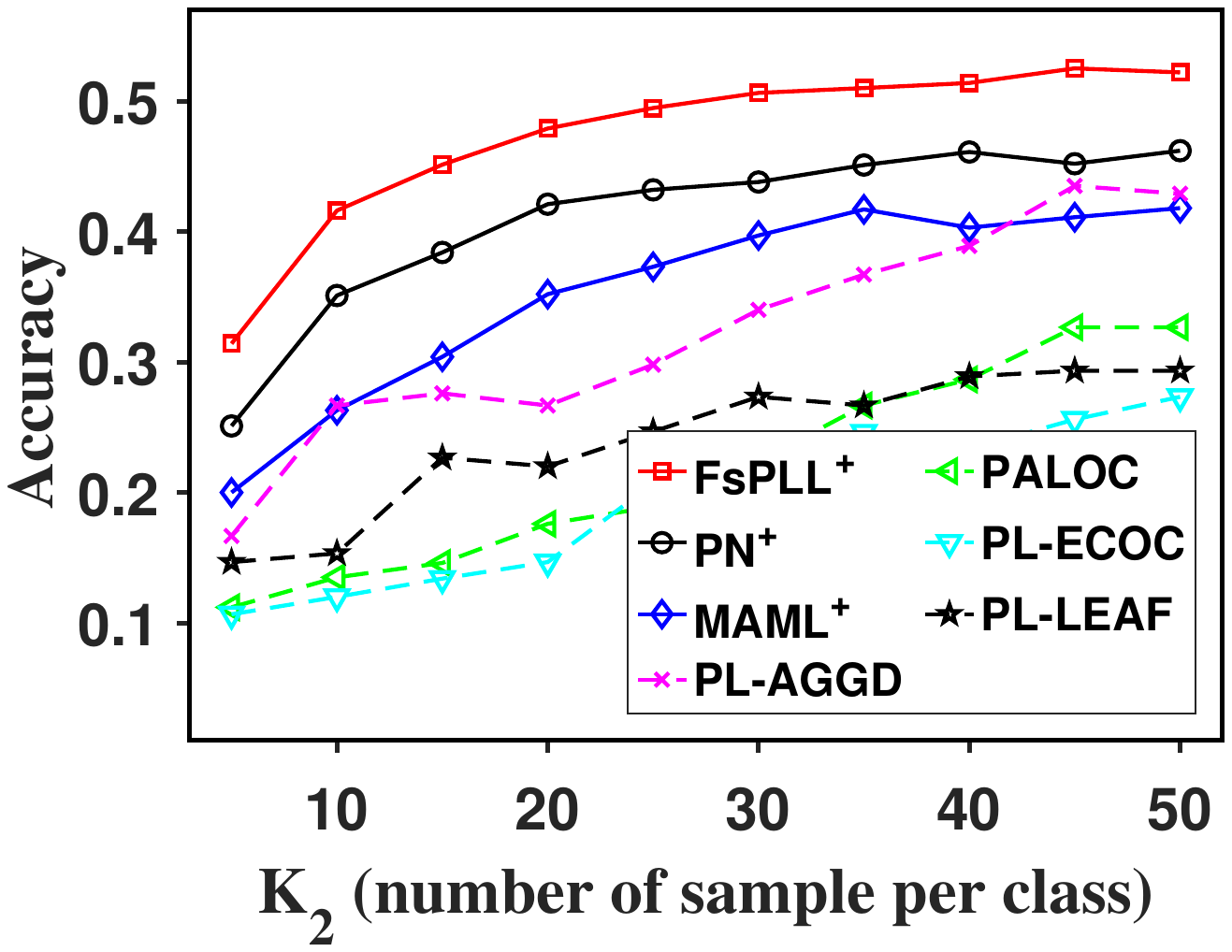}}
\vspace{-1em}
\caption{Accuracy of each compared method vs. $K_2$ (number of support samples per class in the meta-test set) on \textbf{miniImageNet} ($r=1$).  }
\label{miniImageNet_r_1}
\end{figure*}

\noindent\textbf{Compared Methods}: We compare FsPLL against four recent PLL methods (PL-ECOC~\cite{zhang2017disambiguation} , PL-LEAF~\cite{zhang2016partial}, PALOC~\cite{wu2018towards}, PL-AGGD~\cite{wang2019adaptive}), two representative FSL methods (MAML~\cite{finn2017model}, PN \cite{snell2017prototypical}), and FsPLL-nM (a variant of FsPLL) which disregards the local manifold of training samples but updates the label confidence matrix via Eq. \eqref{confidence1} for prototype rectification.
Each compared method is configured with the suggested parameters according to the  corresponding literature, and the configuration details are given in the Supplementary file.
As to our FsPLL, the trade-off parameter $\lambda$ is fixed as 0.5 (0 for FsPLL-nM), the number of nearest neighbors $k=K_2-1$, the number of iterations for prototype rectification in each epoch is fixed to 10, the learning rate is fixed as 0.001 and cut into half per 20 epochs. FsPLL uses the embedding network proposed by~\cite{vinyals2016matching}. For Omniglot, the size of prototypes is $m=64$; while for miniImageNet, $m=1600$. For non-FSL PLL methods, they also used the image features extracted by ~\cite{vinyals2016matching}. They only use the samples in meta-test set  for training and validation. {We randomly generate $\mathcal{D}_{train}$ ($T=100$) as the meta-training tasks in each round, and report average results on  $\mathcal{D}_{test}$ in 100 rounds for reducing the randomness. }

\subsection{Result Analysis}
\textbf{Results on Omniglot}: Table \ref{omniglot} reports the  accuracy of each compared method on \textbf{Omniglot} as $p$ is fixed to 1, $r$ is fixed to 1, 2 or 3, $N_2$ is fixed to 5, 10, 20 or 30, $K_2$ is fixed to 5 or 10. Due to the page limit, we only report the results of each compared method when $r=1$, 2, while the results with $r=3$ are reported in the Supplementary file. From this Table, we have the following observations: \\
(i) FsPLL significantly outperforms other compared methods across all the settings, which proves the effectiveness of FsPLL on few-shot PL samples. The performance margin between FsPLL and non-FSL methods are more prominent w.r.t. a small $K_2$, since these non-FSL methods build on the promise of many-shot PL samples for training. Although PN and MAML additionally use many tasks with support PL samples for the few-shot setting, they often lose to many-shot PLL methods. That is because they are both heavily misled by irrelevant labels of support samples. In contrast, our FsPLL is much less impacted by irrelevant labels of support samples, it reduces the negative impact of irrelevant labels by iteratively rectifying the prototypes and embedding network. By virtue of precise labels of meta-training samples, PN$^+$ and MAML$^+$ outperform many-shot PLL methods, but they still lose to FsPLL$^+$ by a large margin. 
These observations confirm that the noisy labels of PL samples heavily mislead the adaption of meta-learner toward the target task.  \\
 (ii) Prototype rectification can greatly reduce the negative impact of irrelevant labels of PL samples. This is supported by the performance margin between FsPLL  (FsPLL$^+$) and PN (PN$^+$). They both perform distance metric learning in the embedding space to learn prototypes and classify samples therein, but FsPLL additionally rectifies the prototypes in the embedding space  by explicitly modeling irrelevant labels and mining local manifold.\\
(iii) Local manifold helps prototype rectification, this is verified by the clear margin between FsPLL and FsPLL-nM, especially when the number of irrelevant labels is large.\\
(iv) As $N_2$ steps from 5 to 30 under a fixed $K_2$, the performance of each compared method gradually decreases. This is due to the increased class labels and task complexity. The random guess accuracy decrease from 1/5 to 1/30. Even though, FsPLL (FsPLL$^+$) always maintains a better performance than PN (PN$^+$) and MAML (MAML$^+$). On the other hand, as the increase of $K_2$ under a fixed $N_2$, each compared method has an improved performance, since more support samples can be used for training. We see non-FSL PLL methods frequently outperform FSL methods (PN and MAML) when $K_2=10$. This fact again proves the vulnerability of FSL methods on few-shot PL samples.\\
(v) As the increase of $r$, all methods have a reduced performance, since the meta-training PL samples have more irrelevant labels, which seriously compromise the performance of many-shot PLL and FSL methods. This fact signifies the importance to account for PL samples. All  compared methods have a relatively large standard deviation, that is due to the noisy labels of PL samples were randomly injected, and more noisy labels cause an even larger fluctuation. We applied signed-rank test to check the statistical significance between FsPLL/FsPLL$^+$ and other compared methods, all $p$-values are small than 0.001.\\


\vspace{-1em}
\noindent \textbf{Results on miniImageNet}: We also conduct experiments on \textbf{miniImageNet} with the following control setting: $r \in \{1,2,3\}$ with $p=1$, $N_2 \in \{5, 10, 15, 20\}$ and $K_2 \in [5,50]$. We enlarge the range of $K_2$ to check how FsPLL works in the many-shot setting. Due to the page limit, we only report the results of each compared method when $r = 1$. Similar trends can be observed with other settings, which are reported in the Supplementary file. As shown in Fig.~\ref{miniImageNet_r_1}, FsPLL again outperforms state-of-the-art FSL and many-shot PLL methods under different $K_2$ shots, and the conclusions are similar as those on Omniglot. With the increase of $K_2$, all methods show an increased performance, and FsPLL still has a higher accuracy than other methods when $K_2>20$, which proves the effectiveness of FsPLL in many-shot settings.

\subsection{Further Analysis}
\begin{figure}[tb]
\centering
\includegraphics[width=6cm, height=4cm]{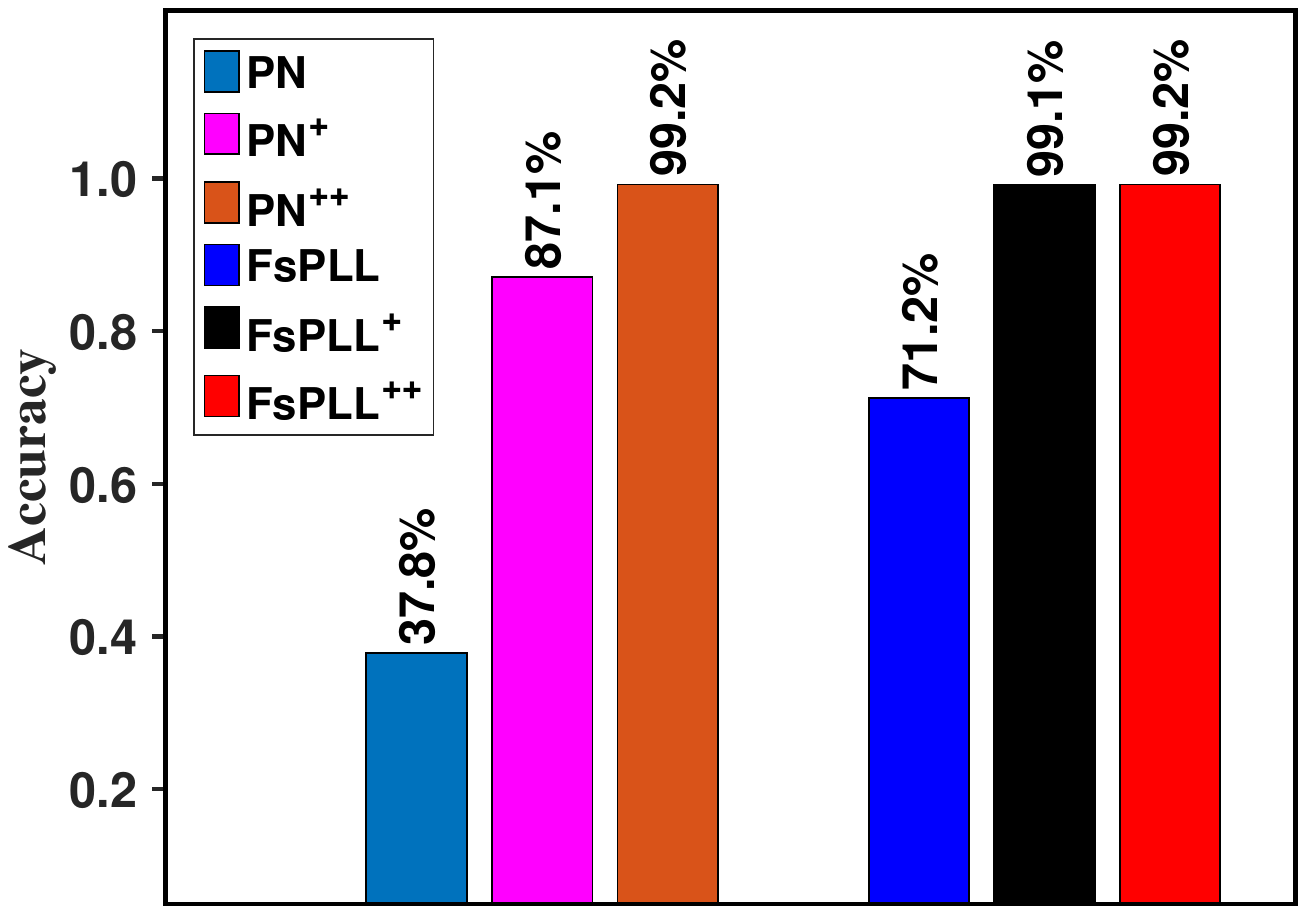}
\vspace{-1em}
\caption{The performance of PN and FsPLL in three different settings on Omniglot. FsPLL$^{++}$ and PN$^{++}$ use the precise labels of meta-training and meta-test samples, give the upper bound accuracy. }
\label{noisy_analyse}
\end{figure}

{\bf Impact of PL samples on FSL methods}: We conduct additional experiments to further investigate the impact of noisy support set of meta-training and meta-test. For this investigation, we introduce another variant FsPLL$^{++}$, which uses precise labels of support samples in the meta-training and meta-test stages. For comparison, we introduce PN$^{++}$ for PN. So  FsPLL$^{++}$/PN$^{++}$ gives the upper bound performance of FsPLL/PN. Fig. \ref{noisy_analyse} shows the performance of FsPLL and PN and their variants under the setting of $N_2 = 10$, $K_2 = 5$ and $r = 2$ on Omniglot. In the figure, FsPLL/PN uses PL samples both in the meta-training and meta-test stages; while FsPLL$^+$/PN$^+$ uses precise labels of support samples in the meta-training stage, and PL samples in the meta-test stage. FsPLL significantly outperforms PN whenever there are support PL samples with irrelevant labels. They can have a comparable performance with precise labels of all support samples. FsPLL improves the accuracy of PN by 88\%, and FsPLL$^+$ improves this of PN$^+$ by 13\%. More importantly, FsPLL$^+$ has a similar accuracy with FsPLL$^{++}$. These results not only confirm the negative impact of noisy PL samples on FSL methods, but also prove the effectiveness of FsPLL on handling noisy labels of PL samples.

\noindent{\bf Parameter analysis}: We also study the parameter sensitivity of FsPLL  w.r.t. $\lambda$ and $k$ (see Eq. \eqref{confidence2}), which uses the local manifold to update the label confidence matrix, and consequently rectify the prototype and embedding network $f_{\theta}$. The results is reported and analyzed in the Supplementary file. We find the local manifold indeed helps rectifying prototypes, $\lambda=0.5$ and $k\approx K_2-1$ give a better performance.

\section{Conclusion}
\label{sec:concl}
This paper studies the problem of few-shot learning with noisy support samples and proves noisy labels of support samples can greatly compromise the performance. We introduce a Few-shot Partial Label Learning approach (FsPLL) to address this problem. FsPLL learns an embedding network and rectifies prototypes to reduce the impact of noisy labels. Extensive experimental results prove the effectiveness of FsPLL in both few-shot and many-shot settings.


\bibliographystyle{named}
\bibliography{FSPLL_Bib}

\end{document}


\maketitle

\section{Experimental Setup}
\subsection{Datasets}
In this section, we conduct experiments on two benchmark FSL datasets (\textbf{Omniglot} and \textbf{miniImageNet}). Following the canonical protocol adopted by previous PLL methods \cite{wang2019adaptive,zhang2016partial}, we generate the semi-synthetic PL datasets on Omniglot and miniImageNet by two controlling parameters $p$ and $r$. Here, $p$ controls the proportion of training examples that are partially labeled, 
and $r$ controls the number of irrelevant labels within candidate label set of each instance. 

\textbf{Omniglot}~\cite{lake2011one} is consisted of 1623 handwritten characters collected from 50 alphabets, each character includes 20 instances drawn by different human subjects. Following~\cite{vinyals2016matching}, we resized the grayscale images to $28\times 28$ and augmented the character classes with rotation in 90, 180 and 270 degrees. We used original 1200 characters plus rotations for meta-training (4800 classes in total) and the remaining classes, including rotations, for meta-test (1692 classes in total). To avoid too repeated computations, each training dataset $\mathcal{D}_{train}^t = \{\mathbf{X}^t_s,\tilde{\mathbf{X}}^t_q,\mathbf{Y}^t\}$ consists of $N_1=30$ randomly sampled classes from 4800 classes without replacement. As to the meta-test set, we randomly selected $N_2$ classes from 1692 classes without replacement. For each selected class, we randomly sampled $K_1=5$ ($K_2$) samples without replacement as training (test) support samples, and took the remaining samples as query samples.  

\textbf{miniImageNet}~\cite{vinyals2016matching} is collected from the larger ILSVRC-12 dataset~\cite{russakovsky2015imagenet} and consisted of 100 classes with 600 examples per class (60000 color images of size $84\times 84$ in total). We use the splits (64 classes for training, 16 classes for validation  and 20 classes for test) proposed by~\cite{ravi2016optimization} for experiments. We use 64 training classes plus 16 validation classes for meta-training, and remaining 20 test classes for meta-test. Similar as what we had set up on \textbf{Omniglot}, each $\mathcal{D}_{train}^t$ consisted of $N_1=30$ classes were randomly sampled from 80 classes without replacement. As to the meta-test set, we randomly selected another $N_2$ classes from 20 test classes without replacement. For each selected class, $K_1=5$ ($K_2$)  samples were randomly chosen from 600 samples without replacement for respective support samples, and another 15 samples per class were randomly chosen as the query samples.

\subsection{Compared Methods} 
To comparatively study the performance of FsPLL, we compare it against four state-of-the-art PLL methods and two representative FSL methods, each configured with the suggested parameters according to the  corresponding literature:
\begin{itemize}
\item \textbf{PL-ECOC}~\cite{zhang2017disambiguation} transforms PLL  into a disambiguation-free problem via error-correcting output codes. Suggested configuration: codeword length $ L = \lceil 10\cdot  \log_2 l \rceil$. 
\item \textbf{PL-LEAF}~\cite{zhang2016partial} learns from PL examples based on feature-aware disambiguation. Suggested configuration: $K = 10$, $C_1 = 10$ and $C_2 = 1$.
\item \textbf{PALOC}~\cite{wu2018towards} adapts one-vs-one decomposition strategy to enable binary decomposition for learning from PLL examples. Suggested configuration: $\mu=10$.
\item \textbf{PL-AGGD}~\cite{wang2019adaptive}  proposes a unified framework to jointly optimize the ground-truth label confidences, instance similarity graph, and model parameters to achieve generalization performance. Suggested configuration: $k = 10,$ $T = 20,$ $\lambda = 1$, $\mu = 1$ and $\gamma = 0.05$.

\item \textbf{MAML}~\cite{finn2017model} is a representative optimization based meta learning algorithm that can adopt to a new task with a small number of gradient steps and a small amount of training data. Suggested configuration: using the same neural network architecture as the embedding function used by~\cite{vinyals2016matching}


\item \textbf{PN} \cite{snell2017prototypical} serves as the baseline of FsPLL, it directly uses PL samples to seek the prototypes.


\item \textbf{FsPLL-nM} disregards the local manifold of training samples to update the label confidence matrix and to rectify the prototypes.

\end{itemize}
As to our FsPLL, the trade-off parameter $\lambda$ is fixed as 0.5 (0 for FsPLL-nM), the number of nearest neighbors $k=K_2-1$, the number of iterations for prototype rectification in each epoch is fixed to 10, the learning rate is fixed as 0.001 and cut into half per 20 epochs. The used embedding network is proposed by~\cite{vinyals2016matching} and is consisted of four convolutional blocks, each of which is  a 64-filter $3\times 3$ convolution followed by a batch normalization layer, a ReLU non-linearity and a $2\times 2$ max-pooling layer. For \textbf{Omniglot}, the size of prototypes is $m=64$, while for \textbf{miniImageNet}, $m=1600$. 

For non-FSL PLL methods, they also used the image features extracted by ~\cite{vinyals2016matching}. They only use the $N_2*K_2$ meta-test support samples for training, and use the same meta-test query samples for validation. {To reduce random impact, we randomly generated $\mathcal{D}_{train}$ ($T=100$) as the meta-training tasks, and $\mathcal{D}_{test}$ as the meta-testing task in each round, report average results in 100 rounds.}

\section{Additional results on Omniglot and miniImageNet}
\begin{figure*}[tb]
\centering
\subfigure[$N_2=5$]{\includegraphics[width=4cm, height=3cm]{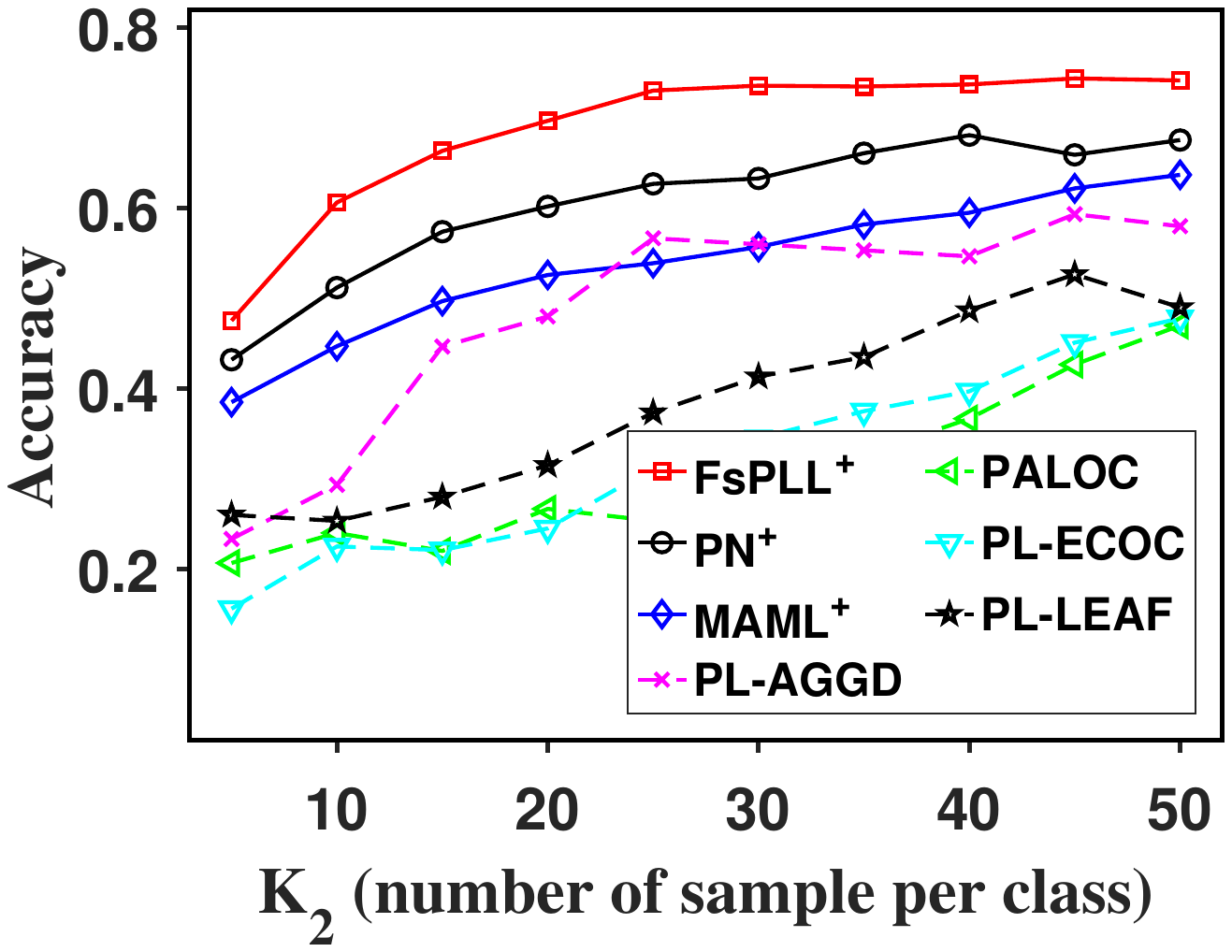}}
\subfigure[$N_2=10$]{\includegraphics[width=4cm, height=3cm]{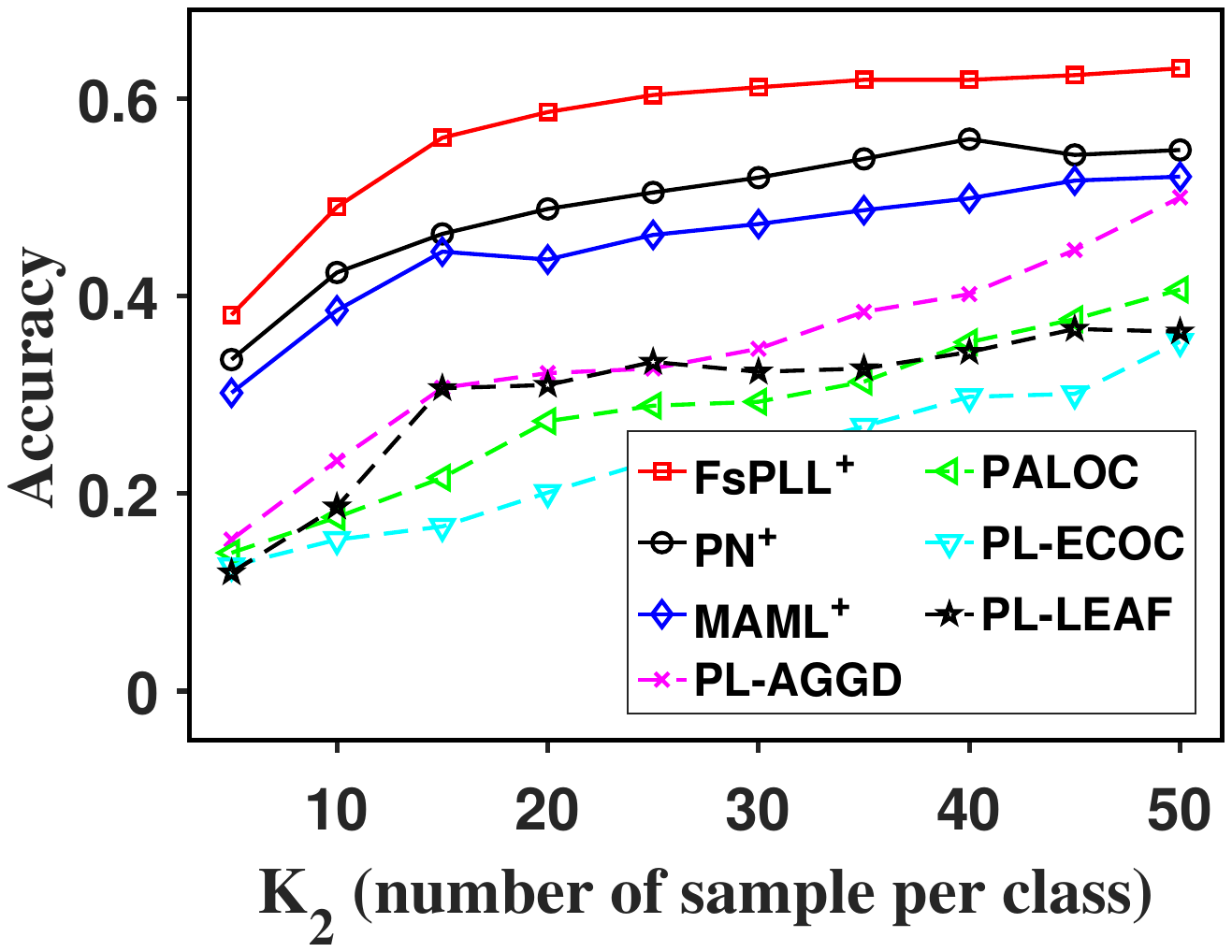}}
\subfigure[$N_2=15$]{\includegraphics[width=4cm, height=3cm]{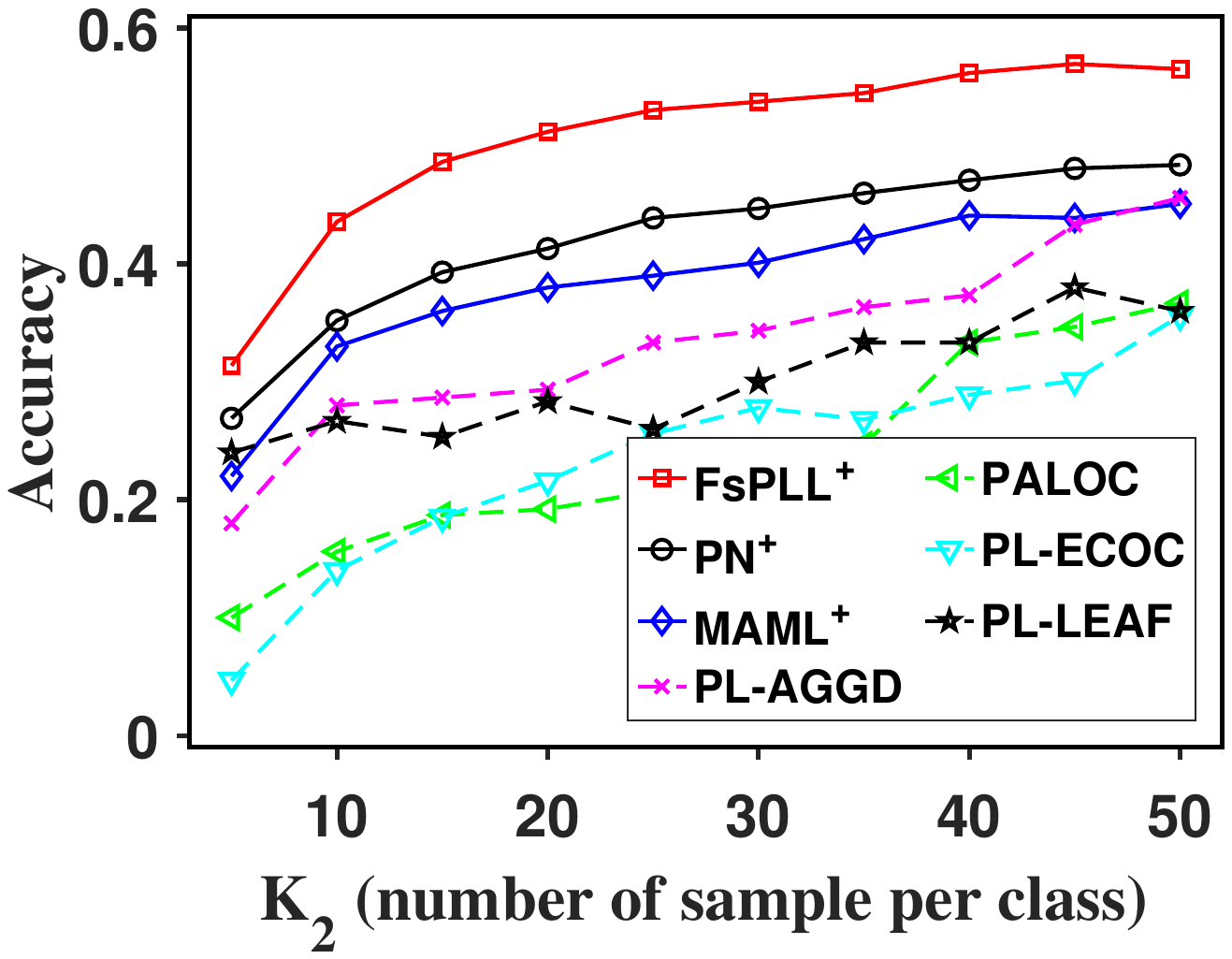}}
\subfigure[$N_2=20$]{\includegraphics[width=4cm, height=3cm]{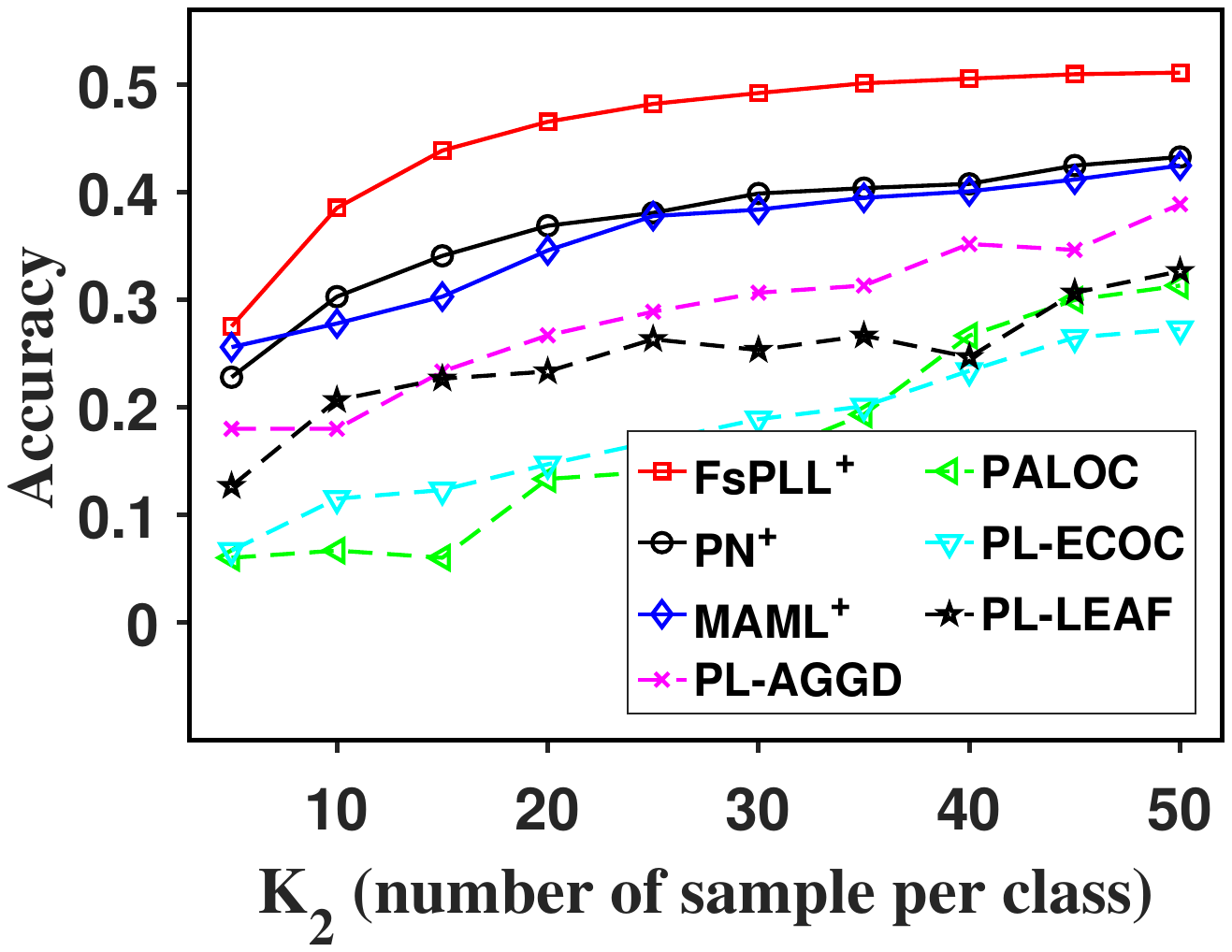}}
\caption{Accuracy of each compared method vs. $K_2$ (number of support samples per class) on \textbf{miniImageNet} ($r=2$).  }
\label{miniImageNet_r_2}
\end{figure*}

\begin{figure*}[tb]
\centering
\subfigure[$N_2=5$]{\includegraphics[width=4cm, height=3cm]{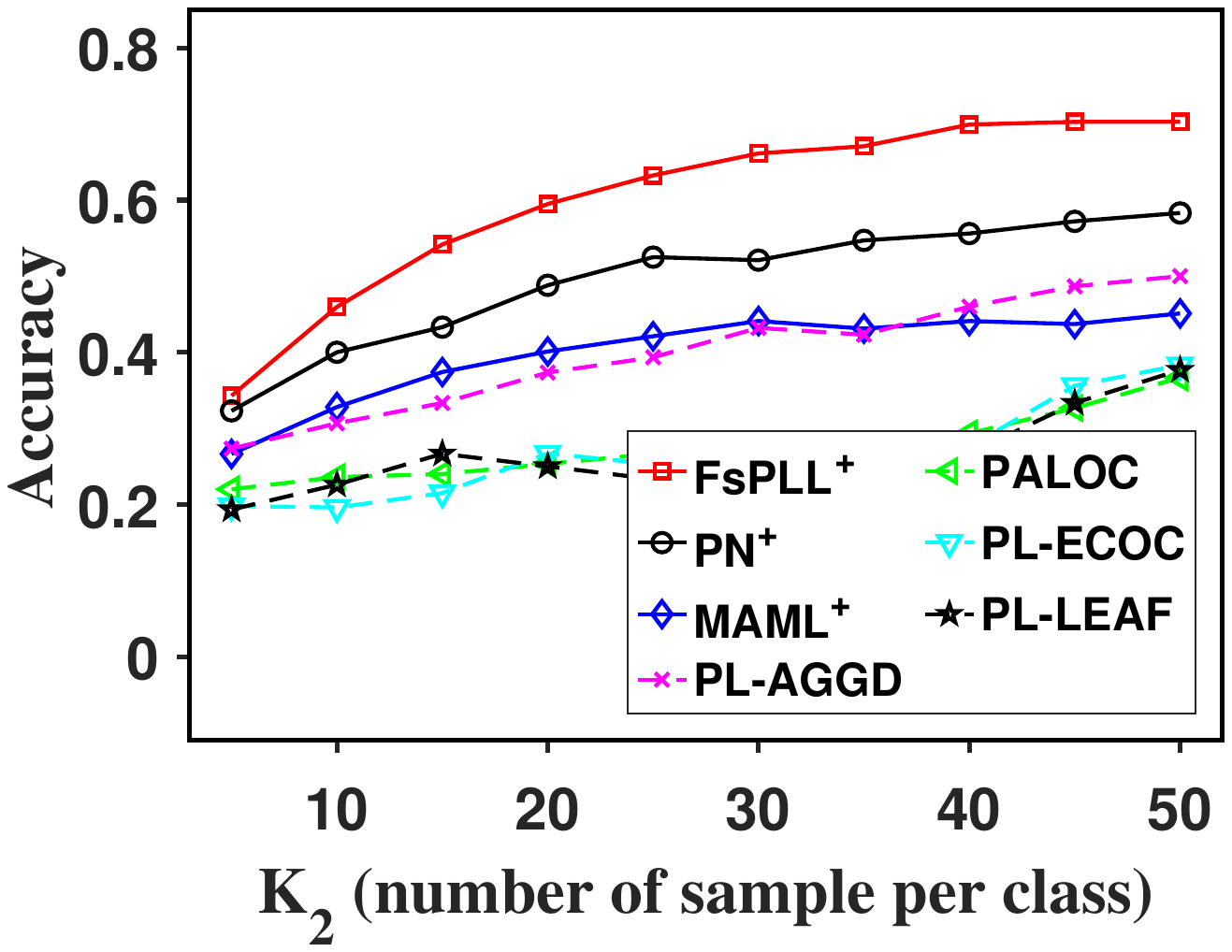}}
\subfigure[$N_2=10$]{\includegraphics[width=4cm, height=3cm]{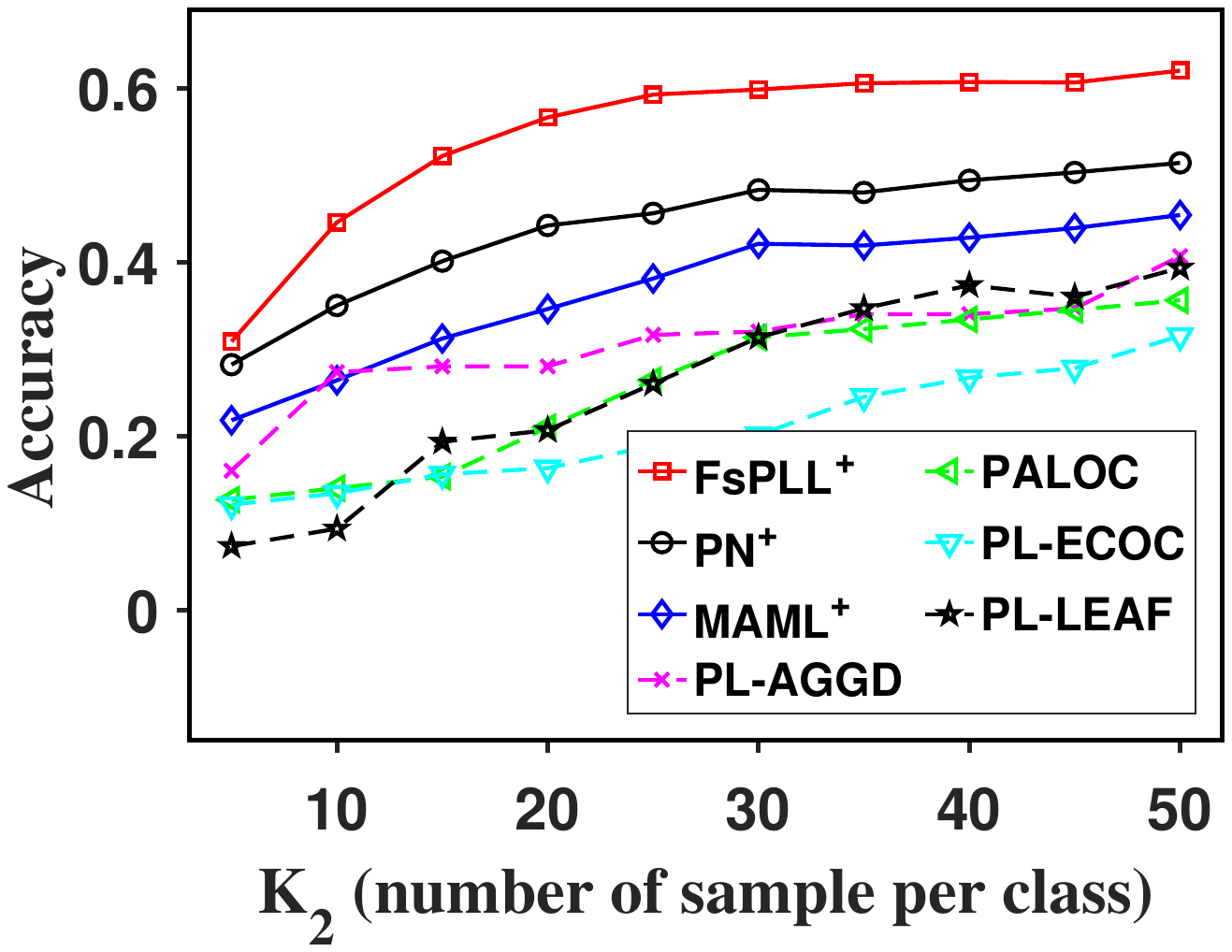}}
\subfigure[$N_2=15$]{\includegraphics[width=4cm, height=3cm]{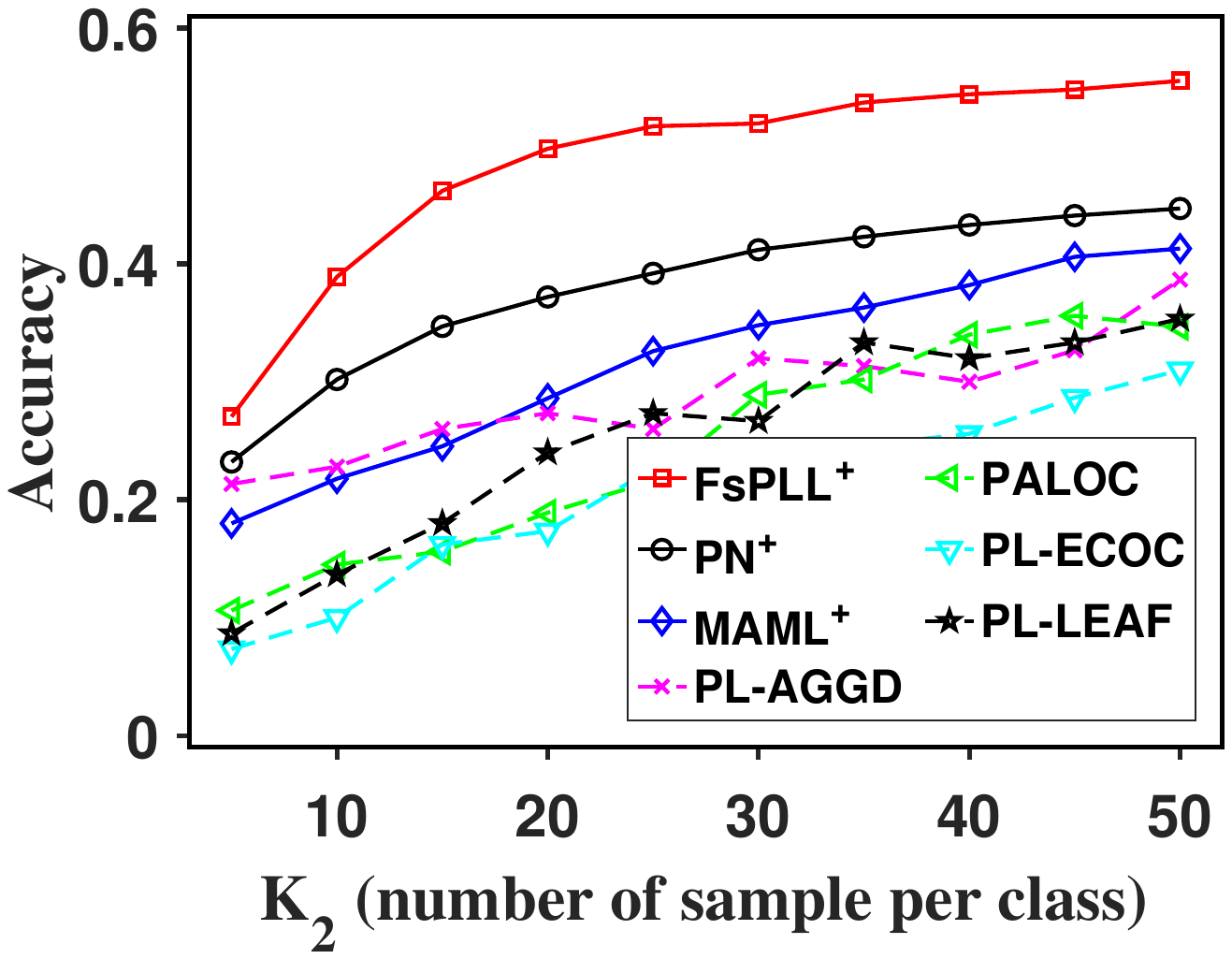}}
\subfigure[$N_2=20$]{\includegraphics[width=4cm, height=3cm]{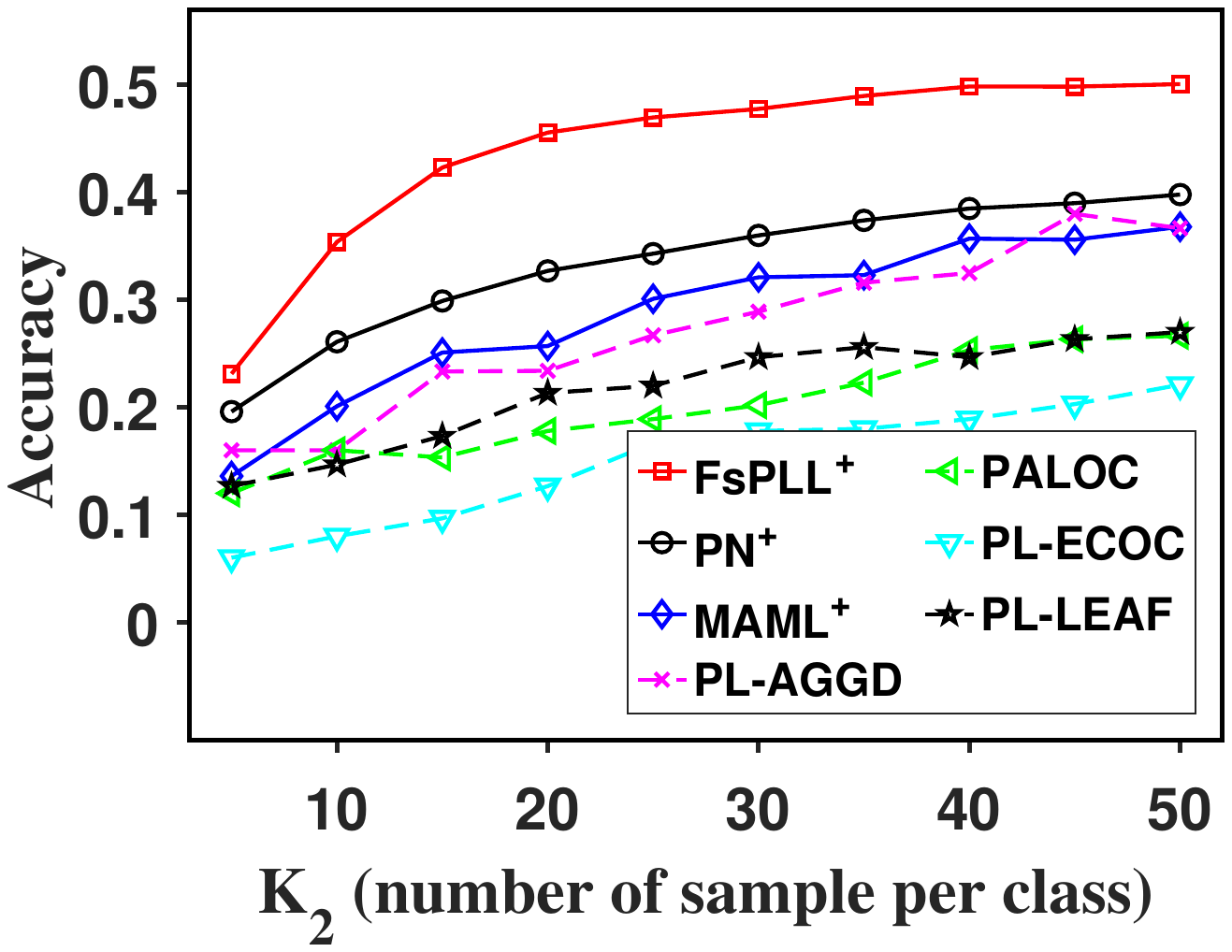}}
\caption{Accuracy of each compared method vs. $K_2$ (number of support samples per class) on \textbf{miniImageNet} ($r=3$).}
\label{miniImageNet_r_3}
\end{figure*}

Table \ref{omniglot} gives the results of compared methods on Omniglot datasets under the setting of $r=3$. Figure \ref{miniImageNet_r_2} and \ref{miniImageNet_r_3} reveal the results of compared methods on miniImageNet with $r=2$ and $r=3$. These results are consistent with the observations stated in the main text, and also prove the effectiveness of our FsPLL.

\begin{figure}[hbt]
\centering
\subfigure[Accuracy vs. $k$]{\includegraphics[width=4cm, height=3cm]{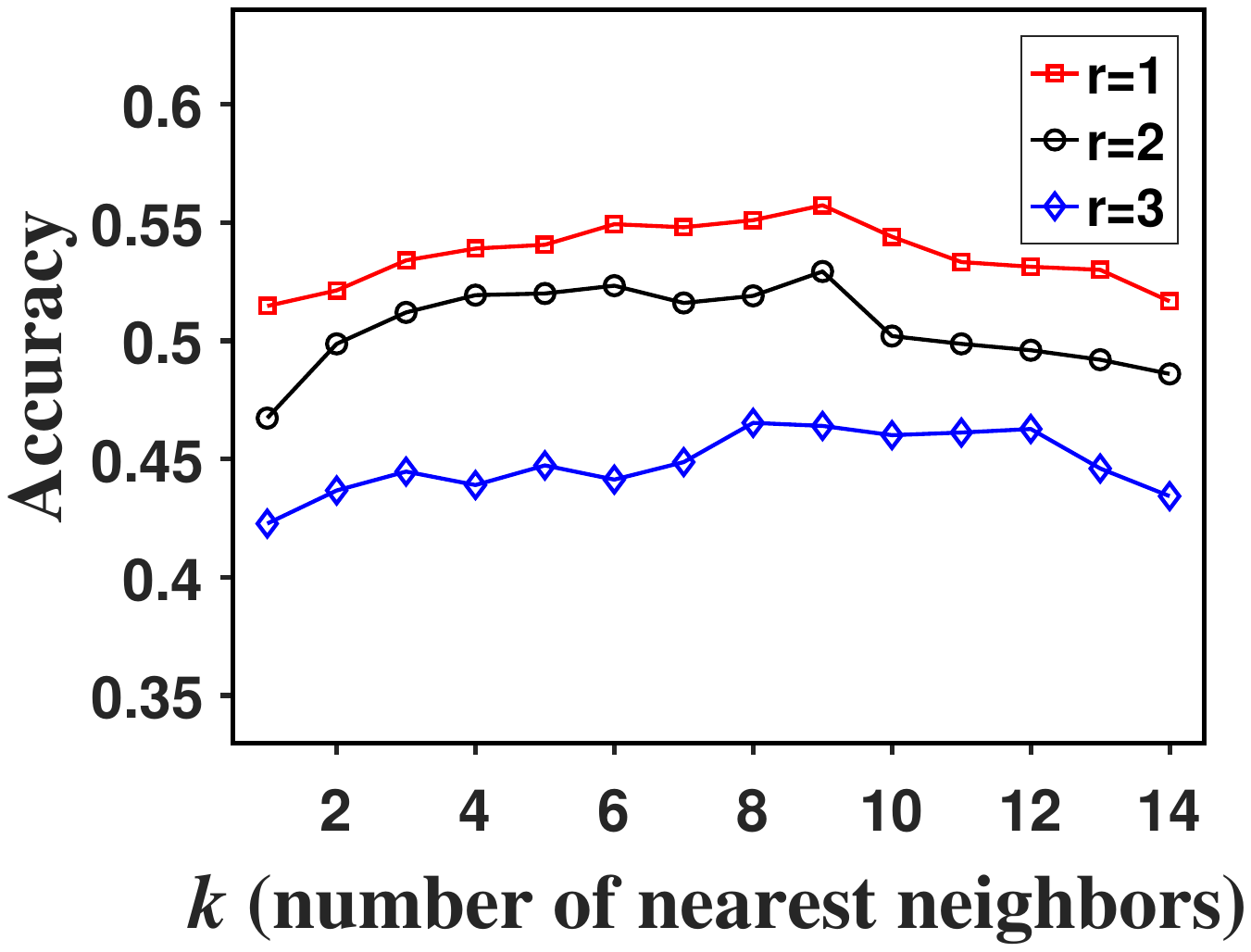}}
\subfigure[Accuarcy vs. $\lambda$]{\includegraphics[width=4cm, height=3cm]{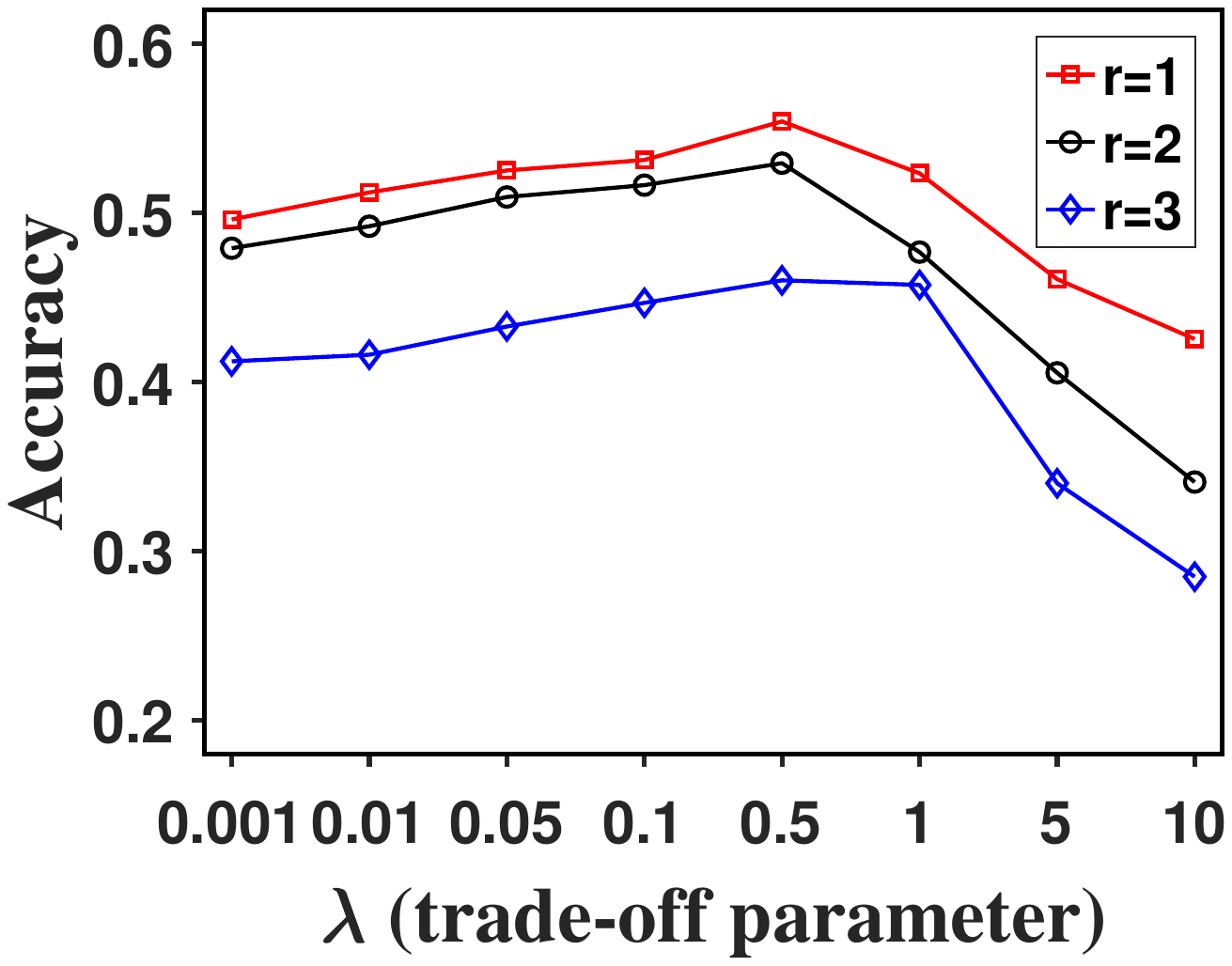}}
\caption{Accuracy of FsPLL on miniImageNet under different input values of $k$ and of $\lambda$, here $N_2 = 10$ and $K_2 = 10$. (a) Accuracy varies with $k$ ($\lambda=0.5$); (b) Accuracy varies with $\lambda$ ($k=K_2 - 1$).}
\label{parameter}
\end{figure}

\begin{table*}[tb]
	\centering
\begin{tabular}{c|ll|ll|ll|ll|}		
		\multicolumn{1}{l|}{} & \multicolumn{2}{c|}{$N_2=5$} & \multicolumn{2}{c|}{$N_2=10$} & \multicolumn{2}{c|}{$N_2=20$} & \multicolumn{2}{c|}{$N_2=30$} \\
		\hline
		\multicolumn{1}{l|}{} & \multicolumn{1}{c}{$K_2=5$} & \multicolumn{1}{c|}{$K_2=10$} & \multicolumn{1}{c}{$K_2=5$} & \multicolumn{1}{c|}{$K_2=10$} & \multicolumn{1}{c}{$K_2=5$} & \multicolumn{1}{c|}{$K_2=10$} & \multicolumn{1}{c}{$K_2=5$} & \multicolumn{1}{c|}{$K_2=10$} \\
		\hline

          & \multicolumn{8}{c|}{$r=3$}                                                                                                                 \\\hline $\text{FsPLL}$ &\textbf{.591± .118} & \textbf{.708±.131} &\textbf{.571±.105} & \textbf{.631±.081} & \textbf{.473±.061} & \textbf{.557±.057}& \textbf{.432±.036} &.\textbf{495±.037}\\
    FsPLL-nM   & .369±.158          & .482±.180          & .404±.109          & .519±.105          & .351±.066          & .461±.064   &  .313±.052 & .415±.050   \\

$\text{PN}$      & .337±.115          & .398±.114          & .277±.066          & .328±.063          & .184±.037          & .222±.037    &.143±.026 & .171±.026\\

$\text{MAML}$       &.379±.124& .469±.112 & .342±.076 & .403±.068 &  .367±.071&  .381±.053 & .345±.023&.407±.061\\

PL-AGGD   & .311±.114          & .439±.134          & .419±.089          & .537±.094          & .376±.068          & .468±.066       &.375±.060 &.453±.054  \\
PALOC     & .308±.117          & .435±.126          & .380±.089          & .523±.092          & .349±.056          & .483±.062       &.310±.051 &   .407±.056\\
PL-ECOC    & .373±.125          & .401±.098          & .293±.076          & .435±.081          & .079±.024          & .264±.048       &.225±.048 & .326±.053  \\
PL-LEAF    & .246±.103          & .452±.109          & .415±.091          & .520±.089          & .388±.056          & .445±.058  & .319±.043& .414±.050 \\\hline

      $\text{FsPLL}^{+}$ & \textbf{.972±.072}          & \textbf{.996±.011}          & \textbf{.990±.014}          & \textbf{.994±.009}          & \textbf{.985±.009}          & \textbf{.989±.012}   &     \textbf{.980±.010}&\textbf{.985±.008}   \\
     
      $\text{PN}^{+}$         & .555±.154          & .722±.143          & .744±.092          & .885±.064          & .748±.062          & .874±.046   &.724±.047 &.852±.037       \\
       $\text{MAML}^{+}$       & .452±.135 & .693±.098 & .608±.076 & .665±.037 &  .473±.027& .598±.035 &.598±.027 &.596±.035\\
          
      \hline     
\end{tabular}
\vspace{-0.5em}
	\caption{Classification accuracy (mean±std) of comparison methods on \textbf{Omniglot}. \{FsPLL, PN, MAML\}$^+$ separately use the precise labels of meta-training support samples. $N_2$: the number of support classes; $K_2$: the number of training samples per class. The best performance in each  setting is \textbf{boldface}.}
\label{omniglot}
\end{table*}

\section{Parameter sensitivity analysis}
Figure \ref{parameter} gives the main text mentioned results of FsPLL under different input values of $k$ and $\lambda$. As shown in Fig.~\ref{parameter}(a) and Fig.~\ref{parameter}(b), FsPLL first manifests a gradually increased accuracy until $k\approx K_2-1$ ($\lambda \approx 0.5$). This trend shows the benefit of local manifold for updating the label confidence matrix and rectifying the prototypes.
However, the accuracy starts to decrease as $k$ and $\lambda$ further increase.  That is due to the over-weight (large $\lambda$) of local manifold and the inclusion of unreliable neighbors (large $k$) from other classes.


\newpage
\bibliographystyle{named}
\bibliography{FSPLL}